\documentclass[journal]{IEEEtran}
\usepackage[T1]{fontenc}        
\usepackage[utf8]{inputenc}     
\usepackage[english]{babel}     

\usepackage{amsmath}            
\usepackage{amssymb}            
\usepackage{mathtools}          
\mathtoolsset{showonlyrefs}     

\usepackage{array}              
\usepackage{tabularx}           
\usepackage{booktabs}           
\usepackage{enumerate}          
\usepackage{enumitem}           

\usepackage{graphicx}           
\usepackage{float}              
\usepackage{caption}            
\usepackage{epstopdf}           

\usepackage{xcolor}             
\usepackage{hyperref}           
\hypersetup{                    
    colorlinks=true, 
    linkcolor=black, 
    citecolor=black, 
    filecolor=black, 
    urlcolor=black
}

\usepackage{listings}           
\usepackage{verbatim}           

\usepackage{mleftright}         
\usepackage{blindtext}          
\usepackage[acronym,nomain]{glossaries}  

\usepackage{tikz}
\usetikzlibrary{arrows.meta, positioning, calc}
\usepackage{adjustbox}
\usepackage{optidef}
\usepackage{algorithm}
\usepackage[noend]{algpseudocode}
\usepackage{subcaption}

\usepackage{filecontents}  
\usepackage[a4paper, total={6.5in, 9in}]{geometry}

\newacronym{sc}{SC}{Semantic Communication}
\newacronym{goc}{GOC}{Goal-Oriented Communication}

\newacronym{bpp}{BPP}{Bits Per Pixel}
\newacronym{bppf}{BPPF}{Bits Per Pixel per Frame}
\newacronym{bpg}{BPG}{Better Portable Graphics}
\newacronym{flif}{FLIF}{Free Lossless Image Format}
\newacronym{jpeg}{JPEG}{}
\newacronym{jpeg2000}{JPEG2000}{}
\newacronym{l1}{L1}{}
\newacronym{l2}{$\ell_2$}{}
\newacronym{fid}{FID}{Frechet Inception Distance}
\newacronym{mse}{MSE}{Mean Squared Error}
\newacronym{mae}{MAE}{Mean Absolute Error}
\newacronym{nmse}{NMSE}{Normalized Mean Squared Error}
\newacronym{psnr}{PSNR}{Peak Signal-to-Noise Ratio}
\newacronym{miou}{mIoU}{mean Intersection over Union}
\newacronym{ssim}{SSIM}{Structural Similarity Index}
\newacronym{lpips}{LPIPS}{Learned Perceptual Image Patch Similarity}
\newacronym{msssim}{MS-SSIM}{Multi-Scale Structural Similarity}

\newacronym{ddpm}{DDPM}{Denoising Diffusion Probabilistic Model}
\newacronym{ddim}{DDIM}{Denoising Diffusion Implicit Model}
\newacronym{nn}{NN}{Neural Network}
\newacronym{dnn}{DNN}{Deep Neural Network}
\newacronym{cnn}{CNN}{Convolutional Neural Network}
\newacronym{ml}{ML}{Machine Learning}
\newacronym{elbo}{ELBO}{Evidence Lower Bound}
\newacronym{gan}{GAN}{Generative Adversarial Network}
\newacronym{vqgan}{VQ-GAN}{Vector Quantized GAN}
\newacronym{sota}{SOTA}{state-of-the-art}
\newacronym{srr}{SRR}{Semantic Relevant Residual}
\newacronym{ssm}{SSM}{Semantic Segmentation Map}
\newacronym{ssmodel}{SSModel}{Semantic Segmentation Model}
\newacronym{sseg}{SSeg}{Semantic Segmentation}
\newacronym{vae}{VAE}{Variational Autoencoder}
\newacronym{vqvae}{VQ-VAE}{Vector Quantized VAE}
\newacronym{maskvqvae}{MQ-VAE}{Masked VQ-VAE}
\newacronym{ae}{AE}{Autoencoder}
\newacronym{unet}{U-Net}{}
\newacronym{resblock}{ResBlock}{Residual Block}
\newacronym{resblockdown}{ResBlock-Down}{}
\newacronym{resblockup}{ResBlock-Up}{}
\newacronym{fc}{FC}{Fully Connected}
\newacronym{mlp}{MLP}{Multilayer Pereptron}
\newacronym{cfg}{CFG}{Classifier-Free Guidance}
\newacronym{pe}{PE}{Positional Embedding}
\newacronym{hific}{HiFiC}{High-Fidelity Generative Image Compression}
\newacronym{fcc}{FCC}{Fidelity-Controllable extreme image Compression}

\newacronym{awgn}{AWGN}{additive white Gaussian noise}
\newacronym{sr}{SR}{Super-Resolution}
\newacronym{lr}{LR}{Low-Resolution}
\newacronym{hr}{HR}{High-Resolution}
\newacronym{roi}{ROI}{Region-Of-Interest}
\newacronym{pca}{PCA}{Principal Component Analysis}
\newacronym{gmm}{GMM}{Gaussian Mixture Model}
\newacronym{rbm}{RBM}{Restricted Boltzmann Machine}

\newacronym{dsslic}{DSSLIC}{Deep Semantic Segmentation based Learned Image Compression}

\newacronym{spic}{SPIC}{Semantic-Preserving Image Coding based on Conditional Diffusion Models}
\newacronym{semcore}{SemCoRe}{Semantic-Conditioned Super-Resolution Diffusion Model}
\newacronym{spade}{SPADE}{Spatially-Adaptive Normalization}
\newacronym{spe}{SemPE}{Semantic Positional Embedding}
\newacronym{amm}{AMM}{Adaptive Mask Module}
\newacronym{samm}{SAMM}{Semantic conditioned Adaptive Mask Module}
\newacronym{adm}{ADM}{Adaptive De-Mask Module}
\newacronym{pi}{PI}{Positional Index}
\newacronym{sqgan}{SQ-GAN}{Semantic Masked VQ-GAN}

\newacronym{dpi}{DPI}{Data Processing Inequality} 
\newacronym{it}{IT}{Information Theory}
\newacronym{ib}{IB}{Information Bottleneck}

\newacronym{en}{EN}{Edge Network}
\newacronym{ed}{ED}{Edge Device}
\newacronym{es}{ES}{Edge Server}
\newacronym{flops}{FLOPS}{FLoating point Operations Per Second}

\newcommand{\x}{\mathbf{x}}
\newcommand{\s}{\mathbf{s}}

\newcommand{\br}{\mathbf{r}}

\newcommand{\Loss}{\mathcal{L}}
\newcommand{\z}{\mathbf{z}}

\newcommand{\y}{\mathbf{y}}
\newcommand{\h}{\mathbf{h}}
\newcommand{\w}{\mathbf{w}}

\newcommand{\e}{\mathbf{e}}
\newcommand{\C}{\mathcal{C}}

\newcommand{\bM}{\mathbf{M}}

\newcommand{\eref}[1]{Eq.~\eqref{#1}}

\newcommand{\fref}[1]{Fig.~\ref{#1}}
\newcommand{\sref}[1]{Section~\ref{#1}}
\newcommand{\cref}[1]{Chapter~\ref{#1}}

\newcommand{\FP}[1]{\textcolor{black}{#1}}
\newcommand{\floor}[1]{\left\lfloor #1 \right\rfloor}

\begin{document}
\title{SQ-GAN: Semantic Image Communications\\ Using Masked Vector Quantization}

\author{Francesco~Pezone{\small{$^{1}$}},
        Sergio~Barbarossa{\small{$^{2}$}},~\IEEEmembership{Fellow ~IEEE}
        and~Giuseppe~Caire{\small{$^{3}$}},~\IEEEmembership{Fellow~IEEE,}
\thanks{$^{1}$ CNIT - National Inter-University Consortium for Telecommunications, Parma, Italy}
\thanks{$^{2}$ Sapienza University of Rome, Italy}
\thanks{$^{3}$ Technical University of  Berlin,  Germany}
\thanks{E-mail: francesco.pezone.ds@gmail.com, \\%
        \hspace*{4.4em}sergio.barbarossa@uniroma1.it, caire@tu-berlin.de}}

\markboth{IEEE Transactions on Cognitive Communications and Networking}%
{Shell \MakeLowercase{\textit{et al.}}: Bare Demo of IEEEtran.cls for IEEE Journals}

\maketitle
\begin{abstract}
This work introduces Semantically Masked \FP{Vector Quantized Generative Adversarial Network} (SQ-GAN), a novel approach integrating \textcolor{black}{semantically driven image coding and vector quantization} to optimize image compression for semantic/task-oriented communications.
\textcolor{black}{The method only acts on source coding and is fully compliant with legacy systems}. \textcolor{black}{The semantics is extracted from the image computing its semantic segmentation map using off-the-shelf software.} A new specifically developed semantic-conditioned adaptive mask module (SAMM) selectively encodes semantically relevant features of the image. \textcolor{black}{The relevance of the different semantic classes is task-specific, and it is incorporated in the training phase by introducing appropriate weights in the loss function.}
SQ-GAN outperforms state-of-the-art image compression schemes 
such as JPEG2000, BPG, and deep-learning based methods across multiple metrics, including perceptual quality and semantic segmentation accuracy on the reconstructed image, at extremely low compression rates. The code is available at \textit{\url{https://github.com/frapez1/SQ-GAN}}.

\end{abstract}


\IEEEpeerreviewmaketitle


\section{Introduction}  \label{intro}
\IEEEPARstart{T}{he} recent surge of intelligent and interconnected devices is leading the transition to new communication paradigms. 
The upcoming sixth generation (6G) of wireless networks is poised to redefine the communication paradigm by integrating advanced Artificial Intelligence (AI) and networking \cite{yang2020artificial}. One of the most promising advancements in this context is semantic communication, which shifts the focus 
\textcolor{black}{from universal coding to task-oriented coding, exploiting suitable knowledge representation systems, typically relying on deep neural networks}
 \cite{Strinati20206G}, \cite{Xie2021Deep, gunduz2022beyond, xie2022task, yang2022semantic}.
Recent breakthroughs in deep learning, generative AI, and self-supervised learning have further strengthened the potential of semantic communications. The use of generative AI for next generation networks has been recently proposed in \cite{barbarossa2023semantic}, \cite{tao2024wireless}, \cite{van2024generative}.
A key innovation in this domain is the integration of deep generative models that can synthesize, with proper conditioning, high-fidelity multimedia contents. This capability changes the communication paradigm: instead of transmitting full-resolution data, \textcolor{black}{the transmitter encodes and sends only the most relevant features, which are used at the receiver side to trigger the generative model to reproduce a representation {\it semantically equivalent} to the transmitted one \cite{tang2024evolving}. The relevance of the features is dictated by the application running on top of the exchange of information. This communication modality enables a drastic reduction in transmission rates, because it takes advantage of application-specific knowledge incorporated in the generative models.} 

To illustrate this concept, consider a vehicular communication scenario, where one or more autonomous vehicles transmit visual information from their onboard cameras to a roadside unit (RSU) responsible for taking real-time context-aware safety decisions, possibly integrating data from multiple vehicles or RSU sensors. 
\textcolor{black}{Rather than transmitting full video streams, each transmitter extracts key semantic attributes, such as the presence, shape, and locations of pedestrians, vehicles, traffic signs, and road boundaries, and transmits only compressed feature representations, achieving data rates substantially lower than those needed for conventional image or video encoding.}
\textcolor{black}{At the receiver side, a generative model reconstructs an application-specific representation, with an accuracy sufficient only to enable the RSU to take rapid decisions within strict latency constraints. }

In recent years, deep learning has played a pivotal role in \FP{both improving image compression and} advancing semantic communication systems \FP{\cite{Xie2021DeepLearningEnabled, mentzer2020hific, Iwai2021fcc}}. The development of end-to-end learning frameworks enables the joint optimization of the encoding and decoding processes, allowing the overall encoding system to learn a joint source/channel coding (JSCC) technique \cite{xu2023deep} or to learn efficient representations of semantic content directly from data \cite{Weng2021Semantic, Xie2021DeepLearningEnabled}. Lately, generative models, \textcolor{black}{such as Variational-Autoencoders (VAE), \glspl{gan}, and Denoising Diffusion Probabilistic Models (DDPM) \cite{goodfellow2014generative}, have been instrumental in improving the quality and efficiency of data representation and transmission \cite{barbarossa2023semantic, tang2024evolving, Liu2024novel, pezone2024semantic}.
}

\textcolor{black}{A key step in the integration of generative models within a digital communication scheme, possibly compliant with legacy systems, is to merge generative models and vector quantization to learn discrete latent representations that are both compact and semantically rich \cite{Oord2017VQ-VAE, Esser2O21Taming}. }
Models like the \gls{vqvae}, the \gls{vqgan} and the \gls{maskvqvae} \cite{Huang2023MaskedVQ-VAE} have already demonstrated remarkable capabilities in image synthesis and compression tasks.

Given this context, in this paper we propose the \gls{sqgan} model, a novel approach that merges the strengths of generative models, vector quantization, and 
semantic segmentation for efficient and semantic-preserving image transmission. \textcolor{black}{We choose the GAN paradigm because it provides a good tradeoff between simplicity and accuracy.} 
\textcolor{black}{In the proposed scheme, semantic information is extracted from the image $\x$ to be transmitted by computing its \gls{ssm} $\s$,
using some suitable state-of-the-art (SOTA) algorithm (in our implementation we used INTERN-2.5 \cite{Wang2022internimage}).}
\textcolor{black}{Then, \gls{sqgan} applies a semantically-guided 
vector quantization strategy to encode both $\x$ and $\s$. 
In particular, \gls{sqgan} learns a vector quantization codebook from the data and applies a newly designed and trained adaptive selective masking to dynamically adjust the compression rate, ensuring that semantically important features are retained.}
This is obtained by \FP{a specifically designed semantic-conditioned variant 
of the} masked vector quantization introduced in \cite{Huang2023MaskedVQ-VAE} 
to prioritize semantically significant information, thus reducing redundancy while preserving the relevant semantic content.
The new adaptive masking mechanism is trained to select the elements in the latent space tensor that are most significant for the preservation of the image's semantic content. Only these selected elements are then quantized, using vector quantization with the learned codebook. Finally, the decoder part of the scheme is formed by some learned blocks, trained to reconstruct the original image by minimizing an error function that includes the \gls{ssm}.
The key innovations in the proposed \gls{sqgan} scheme are: (i) a novel \gls{gan}-based model that integrates semantic conditioning and compression directly into its architecture, (ii) the \gls{samm} that selects and encodes the most semantically relevant features of the input data based on the \gls{ssm}, (iii) a specifically designed data augmentation technique in the training phase 
to enhance the semantically relevant classes, and (iv) the inclusion (in training) of a semantic-aware discriminator to force the model to give more importance to the semantically relevant regions over the non-relevant ones.

\textcolor{black}{Before proceeding, some important remarks regarding our method are in order:}\\ \textcolor{black}{ i) Since we focus on task-oriented semantic communication, we assume that the semantic classes used by the state-of-the-art semantic segmentation algorithm are predefined according to the task requirements. Furthermore, the relevance of each class is quantified by weights \( w_i \in [0, 1] \) assigned to each \(i\)-th class. These weights are incorporated into the loss functions used to train the proposed \gls{sqgan}. The combination of this task-oriented weighted loss function and the aforementioned task-specific data augmentation during training results in an \gls{sqgan} model optimized for the target application. For instance, in our numerical examples focused on autonomous driving scenarios, the relevant object classes include pedestrians, cars, traffic lights, and similar entities.} 

\textcolor{black}{ii) The assumption that the \gls{ssm} $\s$ is extracted from the raw image $\x$ in real-time at the encoder is fully realistic. For example, 
in the autonomous driving scenario, there are algorithms able to extract the SSM in real time, see, e.g. \cite{elhassan2024real}. In particular, Tesla has already deployed the technology to analyze raw images and perform semantic segmentation and object detection in their per-camera networks in real time, as reported in \cite{teslaAI}.} 

\textcolor{black}{iii) Our scheme treats only source coding (i.e., image compression/reconstruction), unlike  several works on semantic communications that have considered a JSCC setting. This makes our approach completely compatible with any underlying legacy network protocol, where source coding/decoding takes place at the application layer and the source-coded bits are transmitted to the 
destination via some standard protocol stack (e.g., a 3GPP wireless network).}

This work is organized as follows. \sref{sec: notation} introduces the scope and general overview of the proposed architecture. \sref{ch: SQGAN} introduces in more detail the different parts of the proposed \gls{sqgan} model. \sref{sec: SQGAN training} discusses the training method with data augmentation for the enhancement of semantically relevant classes and the semantic-aware discriminator employed to optimize the \gls{sqgan}. Finally, \sref{sec: SQGAN numerical results} presents numerical experiments demonstrating the effectiveness of \gls{sqgan} compared to \gls{sota} 
image compression techniques such as \gls{jpeg2000}, \gls{bpg}  \FP{and the \gls{gan} based \gls{hific} \cite{mentzer2020hific} and \gls{fcc} \cite{Iwai2021fcc}}. 

\section{Problem statement and general architecture}\label{sec: notation}

\thispagestyle{plain}
\begin{figure*}[!t]
    \centering
    \includegraphics[width=0.8\textwidth]{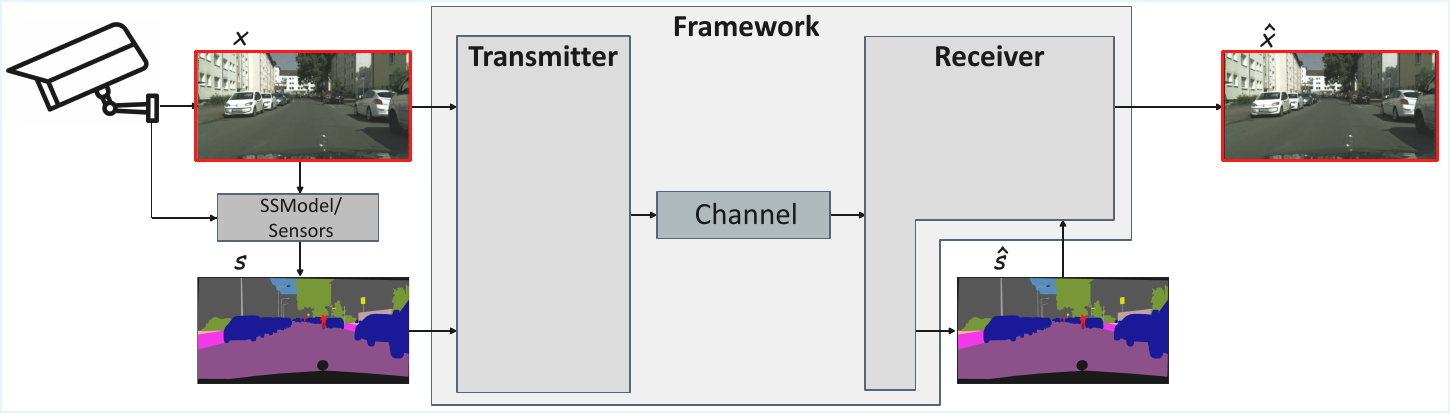}
    \caption[General communication scheme]{Schematic representation of the overall coding/decoding scheme.}
    \label{fig: NOT Communication scheme}
\end{figure*}
\textcolor{black}{The objective of this work is to develop an image encoder that efficiently compresses an image $\mathbf{x}$, while preserving its semantic content as much as possible, even at extremely high compression rates. In this work, 
the semantic content is identified with the \gls{ssm} $\s$, 
extracted in real-time from $\x$ using a \gls{sota} \gls{ssmodel} \cite{Wang2022internimage} with pre-defined semantic classes that depend on the task at hand, as explained in Section \ref{intro}.}

\textcolor{black}{As schematically shown in Fig.~\ref{fig: NOT Communication scheme}, our encoding scheme extracts $\s$ from $\x$ and then jointly encodes $\x$ and $\s$. 
At the decoder, $\x$ and $\s$ are reconstructed as 
$\hat{\x}$ and $\hat{\s}$, respectively. In some sense, this approach is akin layered source coding, where $\s$ can be seen as the ``fundamental layer'' and 
$\x$ is the refinement.  The goal is to achieve good reconstruction performance (with respect to distortion metrics to be defined later) at a low overall coding rate, in terms of  \gls{bpp}. The key result of our work is to show that, even though we transmit two images, namely $\x$ and $\x$ instead of just $\x$, 
we achieve a significantly better trade-off between the overall coding rate and various metric for semantic distortion with respect to SOTA schemes (including data-driven approaches).}

By jointly encoding and decoding $\x$ and $\s$, and incorporating the reconstruction of $\hat{\s}$ in the decoder, 
it is possible to directly optimize the preservation of semantic information {\textcolor{black}{and assign different importance to different semantic classes} during the training process and use $\hat{\s}$ to condition the reconstruction of $\hat{\x}$. Intuitively, this conditioning ``pushes'' the decoder to align $\hat{\x}$
with the semantic structure of the original image $\x$.

\begin{figure*}[!h]
    \centering
    \includegraphics[width=0.8\textwidth]{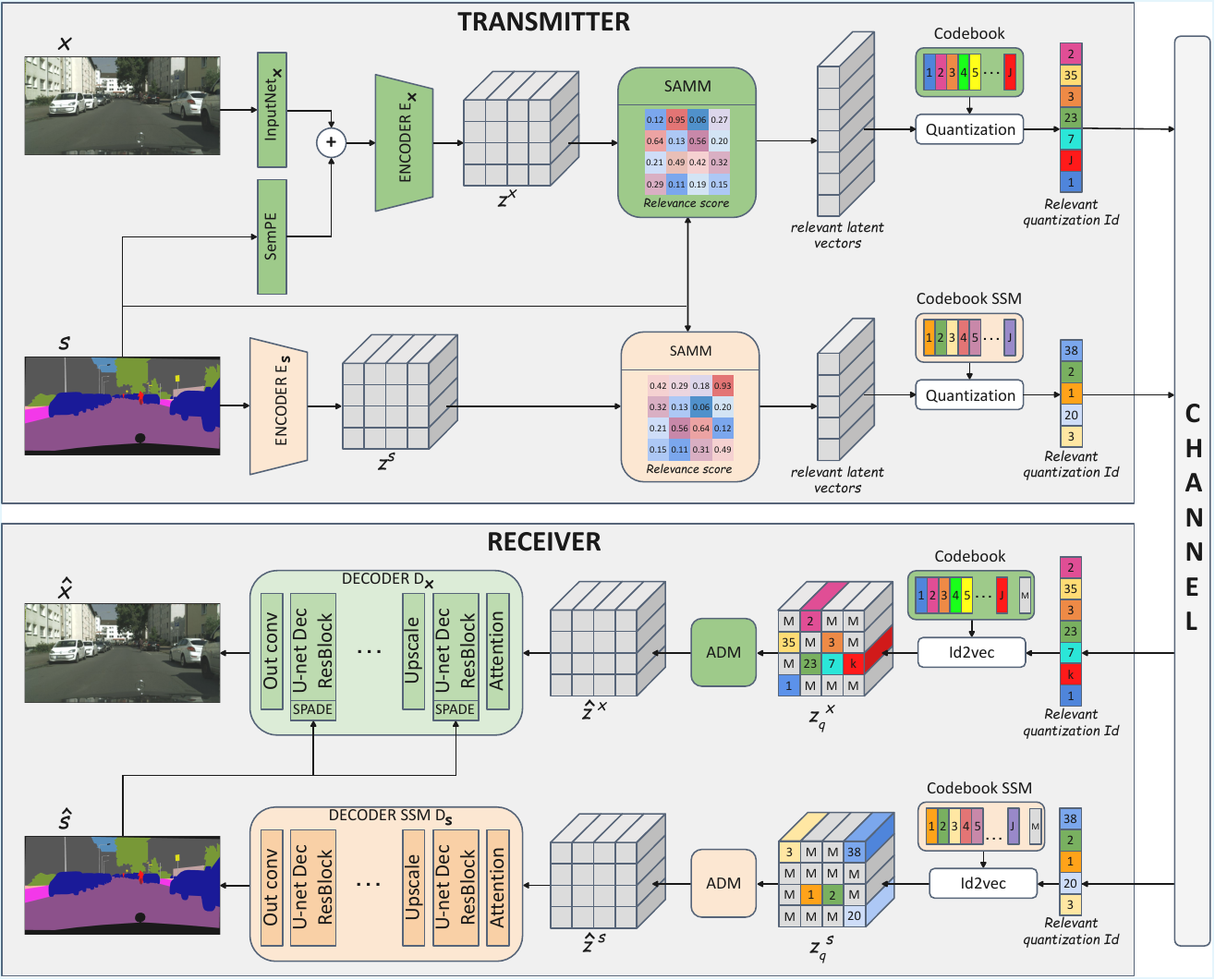}
    \caption[Scheme of the proposed SQ-GAN]{Encoder and decoder detailed structure of the proposed SQ-GAN scheme. The ``channel'' here may represent transmission or storage, depending on the application.}
    \label{fig: SQGAN Scheme masked Sementic VQ-GAN}
\end{figure*}

The detailed architecture of the proposed \gls{sqgan} scheme is shown in \fref{fig: SQGAN Scheme masked Sementic VQ-GAN}, where
the original image $\x$ and the corresponding 
original \gls{ssm} $\s$ are both represented by 3-dimensional tensors. 
The image $\x$ is a tensor of size $3 \times H \times W$, where $3$ refers to the RGB channels, 
$H$ is the height and $W$ is the width of the frame (in pixels).
In this work, we have used images of size $H = 256$ and $W = 512$.
The \gls{ssm} $\s$ is a tensor of size $n_c \times H \times W $, where $n_c$ is the number of semantic classes, $H$ is the height and $W$ is the width of the frame. 
\textcolor{black}{Each $n_c$-dimensional vector in pixel position $(h,w)$ represents the 1-hot encoding of the corresponding semantic class of that pixel. This representation is quite standard and has proven to be convenient in a large number of works on semantic-conditioned image generation.}


\textcolor{black}{The compression and reconstruction of the image $\x$ and of its \gls{ssm} $\s$ is accomplished by two parallel (and mutually interacting) pipelines. We use apex/subscripts $\x$ and $\s$ to indicate blocks of the specific pipeline. Looking at the transmitter side, the bottom pipeline (in beige) reports the encoder of the SSM, whereas the top pipeline (in green) reports the image encoding process, which is conditioned to the SSM. The semantic encoder $E_\s$ takes as input the original \gls{ssm} $\s$ and outputs the latent tensor $\z^\s=E_\s(\s)$ of shape $C \times H_{16} \times W_{16}$ 
where $C=256$, $H_{16}=\frac{H}{16}$ and $W_{16}=\frac{W}{16}$.
In parallel, the image encoding pipeline first extracts the latent representations of $\x$ and $\s$, using two pre-trained DNNs, namely the ${\rm InputNet}_\x$ for $\x$ and the proposed \gls{spe} for $\s$.  These latent representations are added together as $\h(\x, \s) = {\rm InputNet}_\x(\x) + {\rm SemPE}(\s)$ and sent to the encoder that produces the latent tensor $\z^\x = E_\x(\h(\x, \s))$, having the same shape as $\z^\s$.}
Both tensors $\z^\x$ and $\z^\s$  are composed of $K=H_{16} \times W_{16}$ latent vectors $\z_k^\x$ and $\z_k^\s$ 
in $C$ dimensions. 

%

\textcolor{black}{The tensors  $\z^\x$ and $\z^\s$ are then sent to the corresponding \acrfull{samm} blocks, which} select the most semantically relevant latent vectors. \textcolor{black}{The SAMM blocks play a key role in reducing the transmission rate, while preserving the most relevant information.} The number of selected latent vectors depends on a masking fraction variable $m^\x$, or $m^\s$, that can be set as a parameter.  Hence,  
only $N_\x=\floor{m^\x K}$ and $N_\s=\floor{m^\s K}$ selected latent vectors from $\z^\x$ and $\z^\s$ are quantized, respectively. 

Vector quantization (VQ) is applied separately to each selected vector, where the quantization codebook is formed by 
$J$ learnable $C$-dimensional vectors (quantization codewords) denoted by $\e_j^\x$ and $\e_j^\s$, $j \in \{1, \ldots, J\}$, respectively. Therefore, each quantized latent vector is identified by a binary index of 
$\log_2 J$ bits.  In this work we chose $J=1024$ and the \gls{l2} distance as the quantization 
metric (more in \sref{sec: SQGAN quantization}). The quantized latent tensors are denoted by $\z_q^\x$ and $\z_q^\s$, respectively.
The list of quantization indices $\e_j^\x$ and $\e_j^\s$, including the positions of the discarded non-relevant indices, are binary encoded and form the binary encoded stream to be transmitted. 

At the receiver side, the binary stream is decoded and formatted into the latent space tensors $\z_q^\x$ and $\z_q^\s$. Following a similar procedure as in \cite{Huang2023MaskedVQ-VAE}, the latent tensors contain the quantized 
vectors in the positions where these vectors were selected and effectively quantized, and 
a ``placeholder'' codeword $M_\x$ or $M_\s$ in the masked-out positions. These placeholder codewords
are not necessarily all-zero vectors, and can be learned in the training phase in order to facilitate reconstruction. 
These latent space tensors are then processed by the \gls{adm} module to produce the reconstructed latent space tensors $\hat{\z}^\x$ and $\hat{\z}^\s$, respectively.

Finally, the latent tensor $\hat{\z}^\s$ is processed by the decoder $D_\s$ to produce the reconstructed $\hat{s}$ and
the latent tensor $\hat{\z}^\x$ is processed by the decoder $D_\x$ to produce the reconstructed $\hat{x}$ 
with conditioning provided by $\hat{s}$.

\textcolor{black}{The details of the blocks appearing in \fref{fig: SQGAN Scheme masked Sementic VQ-GAN} and the overall training method will be illustrated in the next section.}  
In particular, it is important to clarify that the training of the proposed \gls{sqgan} is performed via an adversarial approach based on 
two discriminators. However, in \fref{fig: SQGAN Scheme masked Sementic VQ-GAN} only the structure of the generator is represented. In fact, the generator is eventually the trained model deployed and used at the run-time, while the discriminators play a role only in the training phase.

\section{Masked Semantic VQ-GAN (SQ-GAN)}\label{ch: SQGAN}
This section describes in detail the individual blocks of the proposed \gls{sqgan} architecture.
\begin{figure*}[!t]
    \centering
    \includegraphics[width=0.75\textwidth]{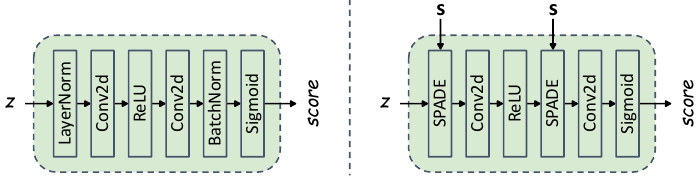}
    \caption[\acrshort{samm} architecture scheme]{Architectural diagram of the \acrshort{amm} as in \cite{Huang2023MaskedVQ-VAE} (left), and the proposed \acrshort{samm} employing the \acrshort{spade} layer to introduce the \acrshort{ssm} conditioning (right).}
    \label{fig: SQGAN SemAdaptiveMask module vs old}
\end{figure*}

\subsection{Semantic Encoder}\label{sec: SQGAN semantic Encoder}
The encoder $E_\s$ maps the $n_c \times H \times W$ \gls{ssm} $\s$ to a latent tensor $\z^\s=E_\s(\s)$ with shape $C \times H_{16} \times W_{16}$, where \textcolor{black}{$C=256$ is the hyperparameter representing the number of channels} and $H_{16}$ and $W_{16}$ are the height and width.
$E_\s$ is formed by a repeated sequence of two \glspl{resblock} \cite{He2016ResBlock} and one down-scaling layer, which will be referred to as \gls{resblockdown}. The average pooling down-scaling layer halves the height $H$ and width $W$ of the \gls{ssm}. 
By applying 4 cascaded stages of \gls{resblockdown}, the final latent tensor has shape $H_{16} \times W_{16}$. 
We decided to use 4 \glspl{resblockdown} stages based on
an extensive ablation study (not reported here because of space limitation), where this choice emerged as the best balance 
between compression and reconstruction accuracy. 
The output of the last \gls{resblockdown} is processed by the final multi-head self-attention layer of the encoder \textcolor{black}{as in \cite{Huang2023MaskedVQ-VAE}}.

\subsection{Image Encoder}\label{sec: SQGAN image Encoder}

The image encoder $E_\x$ maps the image $\x$ to the latent representation $\z^\x$ \textcolor{black}{of shape $C \times H_{16} \times W_{16}$},  using the conditioning provided by $\s$.  $E_\x$ has the same structure as $E_\s$. 
The main difference lies in the conditioning process. In fact, not all the parts of the image $\x$ have the same semantic meaning and relevance: for example, in assisted driving applications, 
pedestrians are more important than the sky. 

To help the encoder $E_\x$ assign different importance to different parts, the image $\x$ is modified before encoding. The proposed method draws inspiration from the version of the \gls{pe} introduced by Dosovitskiy et al. for transformer networks in vision tasks \cite{Dosovitskiy2021ViT}. That scheme divides the image into $16 \times 16$ patches and 
assigns a specific \gls{pe} vector to each patch to allow the transformer to correctly interpret the relative positions of the patches in the frame. In our case, \gls{pe} as proposed in \cite{Dosovitskiy2021ViT} is not needed since,  unlike transformer architectures, the spatial correlations are preserved in convolution-based architectures. However, the idea of assigning different weights to different semantic regions of the image 
is used to improve the overall performance. Hence, we propose a variation of \gls{pe}, referred to as \gls{spe}, whose
goal is to provide the encoder $E_\x$ with a suitable transformation of $\s$ 
that can be used to influence the feature extraction process. \gls{spe} processes $\s$ using a two layer \gls{cnn} designed to take into account the semantic classes of adjacent pixels. 
%
In parallel, the image $\x$ is processed by a one-layer \gls{cnn} called ${\rm InputNet}_\x$. 
The outputs of both \gls{spe} and \gls{cnn} are tensors of shape $128 \times H \times W$.
Then, as in the classical \gls{pe}, the input of the encoder is formed by the elementwise sum ${\h(\x, \s) = \rm InputNet}_\x(\x) + {\rm SemPE}(\s)$. 
This is then processed by the encoder $E_\x$ to produce the latent tensor $\z^\x = E_\x(\h(\x, \s))$ of shape $C \times H_{16} \times W_{16}$. 

\subsection{Semantically Conditioned Adaptive Mask Module}\label{sec: SQGAN SemAdaptiveMask}

The latent tensors $\z^\x$ and $\z^\s$ can be interpreted as collections of $K =H_{16} W_{16}$ 
vectors \textcolor{black}{of dimension $C$}. The next step is to select the most relevant $N_\x=\floor{m_\x K}$ vectors in 
$\z^\x$ and $N_\s=\floor{m_\s K}$ vectors in $\z^\s$, where the relevance masking fraction $m_\x \in (0,1]$ and $m_\s \in (0,1]$ are design parameters.  This selection is performed by the \gls{samm} blocks, which we illustrate next. We only refer to the latent tensor $\z^\x$, as its application to the $\s$ pipeline is similar.

\gls{samm} is a variation of the \gls{amm} introduced in \cite{Huang2023MaskedVQ-VAE}. The key idea consists of assigning to each of the $K$ latent vectors a relevance score and then select the $N_\x$ vectors with the highest score. 
In the proposed \gls{samm}, the relevance score is conditioned on the \gls{ssm} $\s$ and the masking fraction $m_\x$ can be adjusted dynamically. This allows the network to use the same weights to compress images at different levels of compression. The architecture of the conventional \gls{amm} and the new \gls{samm} is shown in \fref{fig: SQGAN SemAdaptiveMask module vs old}.
While the classic \gls{amm}, on the left, is more suitable for general purpose applications, the \gls{samm} on the right is designed to take into account the semantic class of the different regions of the image. The new \gls{samm}  enforces this conditioning thanks to the \gls{spade} normalization layer \cite{Park2019SPADE}. 
\FP{The key advantage of this semantic conditioning is its ability to identify and prioritize the accurate representation of the regions of the image corresponding to the important semantic classes without requiring additional trainable parameters. Without the masking driven by the SSM,
the encoding model would have required many more parameters to learn the features of relevant objects and correctly use this information as input of the \gls{amm}.} 

\gls{samm} takes as input the latent tensor $\z^\x$ and the \gls{ssm} $\s$ and outputs a relevance score $\alpha_k^\x \in [0,1]$ for each latent vector $\z_k^\x$. The final step involves the multiplication of the selected latent vectors by their respective relevance scores. This is done to allow backpropagation to flow through \gls{samm} and train its parameters, as discussed in \cite{Huang2023MaskedVQ-VAE}. 


\subsection{Quantization and Compression}\label{sec: SQGAN quantization}

After selecting the $N_\x$ latent vectors having the highest score, the selected vectors are vector-quantized, while
the not relevant $K-N_\x$ latent vectors are just dropped.
The vector quantization process follows the same steps as the classic \gls{maskvqvae} \cite{Huang2023MaskedVQ-VAE}. 
A learnable codebook $\C_\x = \{ \e_j^\x: j = 0, \ldots, J-1\}$ is used, 
where in this application we found convenient to 
use $J=1024$ codewords of dimension $C=256$.
For each selected score-scaled relevant vector $\z_k^{'\x}= \alpha_k^\x \z_k^\x$, the codeword at minimum distance is found, i.e., the codeword index $j$ is selected such that
\begin{equation}
    j = \underset{i \in \{1, \dots, J\}}{\text{argmin}} \|\z_k^{'\x} - \e_i^\x\|^2,
    \label{eq: GM vq-vae quantization}
\end{equation}
The sequence of quantization indices is then binary encoded using entropy coding.
Every latent vector selected for quantization is located in some position in a $H_{16} \times W_{16}$ shaped array. 
Therefore, the quantized array is formed by $N_\x$ positions containing quantization indices (i.e., integers from 0 to $J-1$ selected as in \eqref{eq: GM vq-vae quantization}) and $K-N_\x$ ``empty'' positions corresponding to the discarded vectors. We assign to all discarded positions an additional special index (conventionally denoted as -1), 
so that we can interpret the quantized array as a discrete information source over an alphabet of size $J+1 = 1025$.
The index -1 appears with probability $1 - m_\x$, while the other indices from 0 to $J-1$ 
appear with probabilities $\beta_j$ such that $\sum_{j=0}^{J-1} \beta_j = m_\x$. 
In general, these probabilities are unknown and depend on the specific image $\x$ to be encoded. 
However, using the fact that entropy is maximized by the uniform probability distribution, we obtain an upper bound to the length of the entropy-coded binary sequence by assuming $\beta_j = \frac{m_\x}{J}$.
Therefore, the entropy coding rate for the index sequence is upper-bounded by:
\begin{equation}
    R_\x = h_2\!\left (m_\x\right) + m_x \log_2(J),
\end{equation}
where $h_2(p) = -(1-p)\log_2(1 - p) - p \log_2(p)$ is the binary entropy function. 
Since the index sequence length is $K$, the number of bits necessary to represent such a sequence is therefore upper-bounded by $B_\x = R_\x K$. Notice that this is indeed a brute-force bound, and in fact any suitable 
more refined compression scheme (e.g., using arithmetic coding coupled with Krichevsky-Trofimov probability sequential estimation \cite{Krichevsky1981universal}) would achieve a lower rate. 
However, in this work this value is further approximated. To maintain a linear relationship between the number of bits and the number of relevant latent vectors, the condition $h_2(p) \leq 1$ is used. 
By substituting $h_2(p)=1$, the final number of bits is expressed as follows:
\begin{equation}
    B_\x = K(1+ m_\x \log_2(J)).
\end{equation}
As will be pointed out in \sref{sec: SQGAN numerical results}, even under this coarse upper bound,  \gls{sqgan} consistently outperforms competing \gls{sota} image compression algorithms. 
By considering the same approach for the $N_s$ relevant vectors in $\z_\s$, $J = 1024$, 
and normalizing by the total number of pixels $H W$, we can express the overall coding rate in \gls{bpp} as a function of $m_\x$ and $m_\s$ as: \footnote{In our results, when referring to \gls{bpp}, it is generally intended this total \gls{bpp}, 
unless otherwise specified.}

\begin{equation}
    {\rm BPP} = \frac{B_\x + B_\s}{H W} = \frac{1}{256}[10 (m_\x + m_\s) + 2],
    \label{eq: SQGAN BPP}
\end{equation}
where $K/(HW) = H_{16} W_{16}/(HW) = 1/256$.
\FP{Notice that this formulation of the BPP considers the contribution of both the  image and its SSM. Therefore, the BPP given above is fully consistent with the definition of source coding rate in image coding (coded bits per image pixel).}
\subsection{Tensor Reconstruction and Adaptive De-Masking Module}\label{sec: SQGAN ADM}

At the receiver, the codebooks $\C_\x$ and $\C_\s$ are stored in the decoder. The bit stream is decoded and the array of 
quantization indices is retrieved. Then, a tensor of shape  $C \times H_{16} \times W_{16}$ is obtained by 
placing quantization codewords $\e_j^\x$ in correspondence of indices $j \in \{0, \ldots, J-1\}$ and a placeholder codeword denoted as $\bM_\x$ in correspondence of all indices equal to -1 (i.e., in the position of the discarded latent space vectors). A similar operation is done for the $\s$ pipeline, where the placeholder is denoted by $\bM_\s$. 
The placeholders are also learned in the training phase. 
\begin{figure*}[!t]
    \centering
    \includegraphics[width=0.9\textwidth]{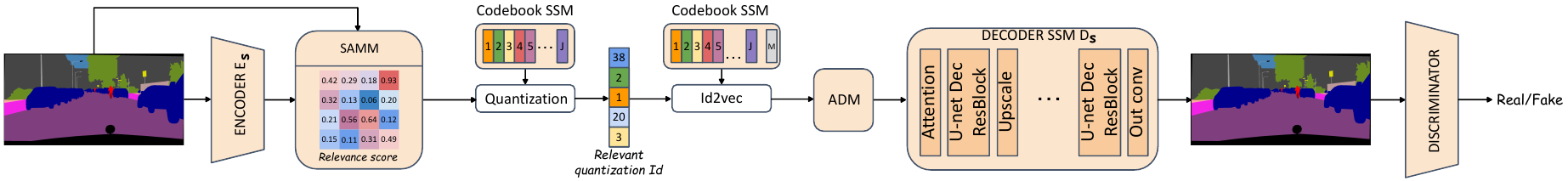}
    \caption[\acrshort{sqgan} training pipeline of the  \acrshort{ssm} generator]{Schematic representation of the semantic generator network $G_\s$ training pipeline.}
    \label{fig: SQGAN Gen_sem sqgan}
\end{figure*}

\begin{figure*}[!t]
    \centering
    \includegraphics[width=0.9\textwidth]{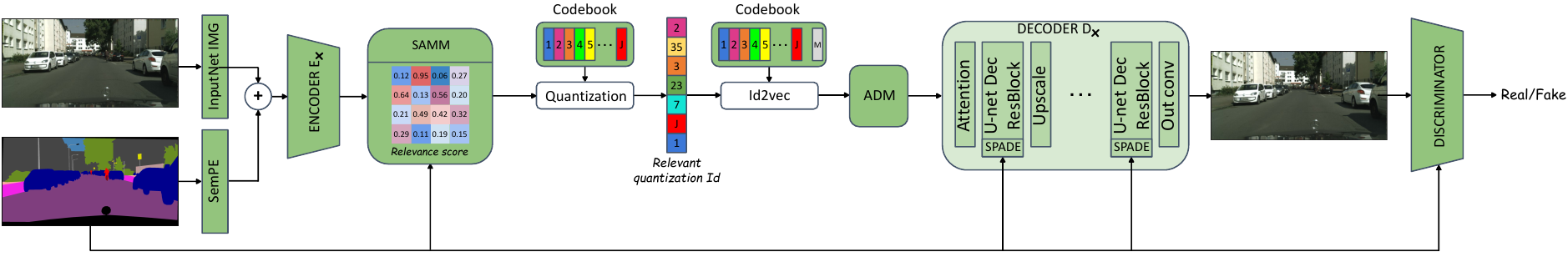}
    \caption[\acrshort{sqgan} training pipeline of the image generator]{Schematic representation of the image generator network $G_\x$ training pipeline.}
    \label{fig: SQGAN Gen_img sqgan}
\end{figure*}

The resulting tensor is processed by the Adaptive De-Masking Module (\gls{adm}). This step is similar to what was introduced in \gls{maskvqvae} \cite{Huang2023MaskedVQ-VAE} and makes use of a direction-constrained self-attention mechanism to gradually let the information flow from the relevant $\e_j^\x$ to the non-relevant placeholder $\bM_\x$. This can be interpreted as a sort of non-linear ``interpolation'' of the latent space tensor in the discarded positions. 
The output $\hat{\z}^\x$ of the \gls{adm} is then used as the input of the decoder $D_\x$. Similarly, the output 
$\hat{\z}^\s$ of the \gls{adm} in the $\s$ pipeline forms the input of the decoder $D_\s$.
At this point, the flows for the two pipelines for $\x$ and for $\s$ diverge again, as described in the next two subsections.

\subsection{Semantic Decoder}\label{sec: SQGAN semantic Decoder}

The structure of the decoder $D_\s$ is similar to a mirrored version of the encoder $E_\s$. The first input layer is composed of the multi-head self-attention layer, after which the series of \glspl{resblockup} is placed. Every \gls{resblockup} is composed of two consecutive \glspl{resblock} and one up-scaling layer. The up-scaling is performed by copying the value of one element in the corresponding up-scaled $2\times2$ patch. The goal is to gradually transform the latent tensor $\hat{\z}^\s$ from a shape of $C\times H_{16} \times W_{16}$ back to the original shape of $n_c \times H \times W$, the same as the \gls{ssm} $\s$. This decoding and up-scaling process is completed by using four consecutive \glspl{resblockup}. 
The final reconstructed \gls{ssm} $\hat{\s}$ is obtained \textcolor{black}{by the standard practice of applying the argmax operator to assign each pixel to a specific semantic class \cite{Ronneberger2015Unet}}. The reconstructed \gls{ssm} $\hat{\s}$ can now be used to condition the reconstruction $\hat{\x}$ of the image $\x$, \textcolor{black}{as described next}.

\subsection{Image Decoder}\label{sec: SQGAN image Decoder}

The reconstruction of $\x$ is obtained by applying the decoder $D_\x$ to $\hat{z}^\x$  and to the reconstructed \gls{ssm} $\hat{\s}$. This decoder has a structure similar to the mirrored version of the encoder $E_\x$, with some key conceptual differences. The input multi-head self-attention layer \textcolor{black}{has the same structure as the one used in the semantic decoder}. The sequence of four consecutive \gls{resblockup} is modified to incorporate the conditioning via the \gls{ssm}. The modifications focus on the normalization layers within the \glspl{resblock}, replacing every normalization layer with the \gls{spade} layer. This new layer is responsible for enforcing the structure of the \gls{ssm} during the decoding phase \cite{Park2019SPADE}.

\section{Training and Inference}\label{sec: SQGAN training}

\textcolor{black}{So far we have provided a structural and functional description of the blocks that compose the encoder and the decoder, as shown in \fref{fig: SQGAN Scheme masked Sementic VQ-GAN}.
In this section, we illustrate the training phase of all these blocks, considering both image and  \gls{ssm} pipelines. 
\textcolor{black}{We recall that the GANs rely on the interplay between a generator and a discriminator. The generator $G_\s$ of the semantic part is coupled with the corresponding discriminator, denoted as $D_{disc}^\s$, while the image generator $G_\x$ is paired to the corresponding discriminator $D_{disc}^\x$.} \FP{The discriminator is a neural network that receives either an image or the \gls{ssm} as input and assesses its authenticity by distinguishing between real and generated samples. It is used exclusively during training to guide the generator, providing feedback on how realistic its outputs are. This feedback enables the generator to progressively improve the quality and realism of the data it produces.} The training scheme for $G_\s$ is shown in Fig.~\ref{fig: SQGAN Gen_sem sqgan}, while that for $G_\x$ is shown in Fig.~\ref{fig: SQGAN Gen_img sqgan}. At runtime (i.e., when the trained model is applied to a generic image for compression and reconstruction), the discriminators play no role.}

\begin{figure*}[!t]
    \centering
    \adjustbox{valign=t}{\includegraphics[width=0.45\textwidth]{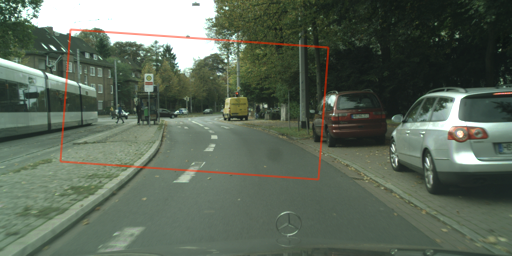}}%
    \hspace{5mm}
    \adjustbox{valign=t}{\includegraphics[width=0.45\textwidth]{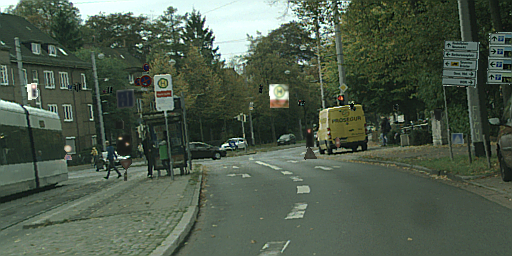}}\\[1mm]
    \adjustbox{valign=t}{\includegraphics[width=0.45\textwidth]{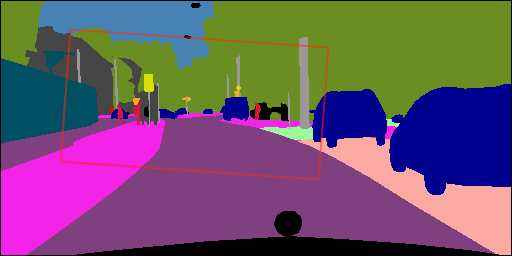}}%
    \hspace{5mm}
    \adjustbox{valign=t}{\includegraphics[width=0.45\textwidth]{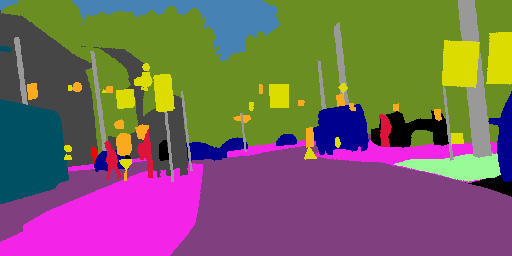}}
    \caption[Data Augmentation for \acrshort{sqgan}]{Effect of combined data augmentation techniques, including rotation, cropping and the proposed semantic relevant classes enhancement. On the left, the original image and \gls{ssm} and on the right the resulting augmented version.}
    \label{fig: SQGAN Data augmentation}
\end{figure*}

The two pipelines are intended to reconstruct objects in distinct domains -- the image domain and the \gls{ssm} domain -- each requiring different loss functions. 
After intensive numerical experimentation, we realized that
training the entire network as a monolithic entity (i.e., of both pipelines coupled together as in Fig.~\ref{fig: SQGAN Scheme masked Sementic VQ-GAN})  is not efficient. 
Additionally, it is useful to consider the masking fractions $m_\s$ and $m_\x$ as variable rather than fixed as in \cite{Huang2023MaskedVQ-VAE}. This allows \gls{sqgan} to compress images and \glspl{ssm} at various compression levels, but also increases the training complexity. 
Hence, to effectively train \gls{sqgan}, we devised a multi-step approach consisting of three stages: (i) train $G_\s$ using the original $\s$, (ii) train $G_\x$ with the original $\x$ and $\s$ and (iii) fine-tune the entire network, denoted by $G$, using the original $\x$, original $\s$, and the reconstructed $\hat{\s}$ by freezing $G_\s$'s parameters and only fine-tuning $G_\x$'s parameters. 
Initially training $G_\s$ ensures that the \gls{ssm} is accurately reconstructed independently of the rest. At the same time, training $G_\x$ with the original $\x$ and $\s$ allows learning the conditional dependencies required for image reconstruction based on a reliable \gls{ssm}. However, since the decoder part in $G_\x$ is later required to be conditioned by the reconstructed $\hat{\s}$ rather than the original $\s$, the fine-tuning step is crucial to adapt $G_\x$ to mitigate the imperfections of $\hat{\s}$ that are not present in $\s$.

In addition to the multi-step training approach, another crucial challenge is the identification and preservation of semantically relevant information. To address this problem, the training process has been further improved by proposing and incorporating a data augmentation step, adding semantically relevant classes, and a {\em semantic-aware discriminator network}. The idea is to emphasize the importance of semantically relevant classes. For example, in our numerical experiment we focused on dashboard camera views for assisted driving applications, where the semantic classes of interest are 
(for example) "traffic signs", "traffic lights", and  "pedestrians". 
These objects typically occupy a small portion of the frame and are not always present in every frame.
In contrast, non-relevant classes such as "sky", "vegetation", and "street", appear much more frequently and occupy a large portion of the images. Therefore, Data augmentation is used to ensure that frequent but task-irrelevant semantic classes do not dominate the training process.

This section is organized as follows. \sref{sec: SQGAN Data Augmentation} will present the data augmentation process as well as the proposed method for enhancing semantically relevant classes. \sref{sec: SQGAN training G_s} will focus on the training of $G_\s$, while \sref{sec: SQGAN training G_x} will focus on the training of $G_\x$ with the introduction of the proposed semantic-aware discriminator. Finally, \sref{sec: SQGAN training G} will focus on the final fine-tuning process. 
\textcolor{black}{The dataset used for training is the Cityscapes dataset \cite{Cordts2016Cityscapes} composed of 2975 pairs of images and associated SSMs.}

\subsection{Data Augmentation}\label{sec: SQGAN Data Augmentation}

This section introduces a novel data augmentation technique aimed at addressing the issue of underrepresented but semantically important classes in datasets.
This new data augmentation technique is fundamentally different from previously proposed methods (e.g., \cite{Konushin2021dataAug1, Jockel2021dataAug2}). These approaches often focus on swapping existing objects -- for example, replacing a "turn left" sign with a "stop" sign -- or employ complex \gls{nn} architectures to introduce new objects into images. These approaches can lead to increased computational costs or fail to adequately address the under-representation of critical classes. In contrast, the proposed technique is straightforward, efficient, and fast.

For the sake of illustration, let us focus on data augmentation to address the under-representation of "traffic signs" and "traffic lights". The process is based on the use of mini-batches of pairs $(\x,\s)$.
For each image in a mini-batch, the \gls{ssm} is used to identify all instances of "traffic signs" and "traffic lights" present in that image. Then, each image and its corresponding \gls{ssm} is augmented 
by adding more instances of these critical classes. This is achieved by copying "traffic signs" and "traffic lights" from other images within the same mini-batch and pasting them into the current image and \gls{ssm}. By increasing the presence of these objects, the model will be more exposed to them during training.

The process begins by collecting all "traffic signs" and "traffic lights" from the images and \glspl{ssm} in the current mini-batch. For each image in the mini-batch, a random number $n$ between $0$ and $25$ is selected, representing the number of objects to add. Then, $n$ objects are randomly selected from the collected set and carefully inserted into the image and its \gls{ssm}. The placement is carried out in such a way as to avoid overlapping with existing instances of the same class or other semantically relevant classes.
 An example of an augmented pair $(\x,\s)$ is represented in \fref{fig: SQGAN Data augmentation} with the original pair on the left and the data augmented version on the right. Other classic data augmentation techniques like cropping, rotation and color correction are also applied. 
However, it is interesting to notice the increase in "traffic signs" (yellow), and "traffic lights" (orange) in the augmented $\x$ and $\s$.

\subsection{Training $G_\s$}\label{sec: SQGAN training G_s}
 
The training of the sub-network $G_\s$ is implemented as depicted in \fref{fig: SQGAN Gen_sem sqgan}, based on an adversarial approach \glspl{vqgan} applied to the training set with data augmentation.
The input and output tensors $\s$ and $\hat{\s}$ of $G_\s$ have a shape $n_c \times H \times W$. We considered the following loss function:
\begin{align}
    \Loss_{\text{SQ-GAN}}^\s &= \FP{\Loss_{\text{WCE}}  + } \\
    & \FP{\hspace{-4mm} + \lambda_{\text{GAN}} \left[ \log D_{disc}^\s(\s) + \log \left( 1 - D_{disc}^\s(G_{\s}(\s)) \right) \right]} \\
    & \FP{\hspace{-4mm} + \lambda_{\text{vq}} \|\text{sg}[\z^{'\s}] - \z^{\s}_q\|_2^2 + \lambda_{\text{commit}} \|\z^{'\s} - \text{sg}[\z^{\s}_q]\|_2^2}\\
    &\hspace{-6mm} := \Loss_{\text{WCE}}  + \lambda_{\text{GAN}} \Loss_{\text{GAN}} + \lambda_{\text{vq}} \Loss_{\text{vq}} + \lambda_{\text{commit}} \Loss_{\text{commit}}   \label{Total semantic loss}
\end{align}
\FP{where the $\lambda$'s are hyperparameters that control the strength of each term. The last three terms in \eqref{Total semantic loss} are the same as in  \cite{Esser2O21Taming} for the conventional \gls{vqgan}. 
More specifically, $\Loss_{\text{GAN}}$ is the term responsible for adversarial training. This term encourages the generator to produce outputs that are indistinguishable from real semantic maps by penalizing the generator if the discriminator can differentiate between the real and generated images.} The structure of the discriminator $D_{disc}^\s$ is that of a neural network as described in \cite{isola2017image2image}. This is composed of a convolutional layer, batch normalization layer and a leaky ReLU \cite{Bing2015Rectified} repeated 3 times and followed by the last convolutional 
layer that maps the output to a single number. This value is the output of the discriminator used to classify the \gls{ssm} as real or fake. \FP{As a result, $\Loss_{\text{GAN}}$ improves the overall quality of the generated semantic reconstructions by guiding the model toward realistic results. 
The $\Loss_{\text{vq}}$ term is responsible for training the codebook used in the quantization process. It utilizes the $\text{sg}[\cdot]$, which is the stop-gradient operator. This operator prevents the gradients from being backpropagated to the encoder during training. By blocking the backpropagation, the encoder's parameters are not updated, and the codebook is encouraged to adjust its codewords to better match the output of the encoder.
The $\Loss_{\text{commit}}$ term ensures that the encoder does not overfit to only a small subset of the codebook entries. This is obtained by penalizing the difference between the encoder's output $\z^{\prime}$ and the quantized codebook entry $\text{sg}[\z_q]$, thus promoting 
the encoder to use a more diverse set of codebook entries. As a result, the generalization and stability of the training process are improved while ensuring that the codebook entries are not updated during the training of the encoder.}

\FP{The first term of the loss function in \eref{Total semantic loss} is a novelty introduced in this work} and consists of the weighted cross-entropy $\Loss_{\text{WCE}}$ defined as: 
\begin{equation} 
\label{Loss_WCE}
    \Loss_{\text{WCE}}(\s, \hat{\s}) = - \sum_{(h,w)} \text{w}_{\s_{(h,w)}} \s_{(h,w)} \log\left( \s_{(h,w)} \cdot \hat{\s}_{(h,w)} \right),
\end{equation}
where $\s_{(h,w)}$ and $\hat{\s}{(h,w)}$ denote the one-hot encoded vectors at pixel location $(h,w)$
in $\s$ and $\hat{\s}$, respectively, and $\text{w}_{\s_{(h,w)}} \geq 0$ is the weight associated with the semantic class in that pixel. 
This choice is motivated by the nature of the \gls{ssm}, which is a pixel-wise classification map where each pixel belongs to a specific semantic class. 
The weights are assigned to emphasize the importance of semantically relevant classes over nonrelevant ones, encouraging the model to focus more on reconstructing the critical classes. Specifically, in the training of the model used in our numerical examples, these 
weights are set as follows:

\vspace{0.2cm}
\begin{tabular}{ll}
    \textbullet \;\; $\text{w} = 1$  & for the relevant classes "traffic signs" \\ &  and "traffic lights". \\
    \textbullet \;\; $\text{w} = 0.85$  & for the classes "people" and "rider". \\
    \textbullet \;\; $\text{w} = 0.20$  & for the non-relevant classes "sky" \\ &  and "vegetation". \\
    \textbullet \;\; $\text{w} = 0.50$  & for all the other classes.
\end{tabular}\\

After defining the loss function, the subnetwork $G_\s$ and the associate discriminator $D_{disc}^\s$, can be trained. During training, the masking fraction $m_\s$ is randomly varied, selected from a set of values ranging from 5\% to 100\% with an expected value of 35\%. This approach allows the model to learn to compress the \gls{ssm} at various levels of compression \FP{and increase the robustness of the model with respect to different conditions of the channel. If the task will require it, it is in fact possible to dynamically adapt the masking fraction without having to train another model from scratch, like other approaches like \cite{mentzer2020hific, Iwai2021fcc} and \cite{Huang2023MaskedVQ-VAE} do}.
The subnetwork $G_\s$, and the associated discriminator $D_{disc}^\s$, are trained using the Adam optimizer \cite{Kingma2015Adam} with a learning rate of $10^{-4}$ and a batch size of 8. The training is conducted for 200 epochs with early stopping to prevent overfitting. 

\subsection{Training $G_\x$}\label{sec: SQGAN training G_x}

The training of the subnetwork $G_\x$ follows a similar approach. As said before, at this stage 
$G_\x$ is trained using the original image $\x$ and the original \gls{ssm} $\s$, as illustrated in \fref{fig: SQGAN Gen_img sqgan}.
Again, an adversarial approach typical of \glspl{vqgan} is employed for training, with the loss function defined as:
\begin{align}
    \Loss_{\text{SQ-GAN}}^\x &= \FP{\Loss_{Wl_2}  + \Loss_{\text{perc}} +} \\
    & \FP{\hspace{-6mm} + \lambda_{\text{GAN}} \left[ \log D_{disc}^\x(\x|\s) + \log \left( 1 - D_{disc}^\x(G_{\x}(\x)|\s) \right) \right]} \\
    & \FP{\hspace{-6mm} + \lambda_{\text{vq}} \|\text{sg}[\z^{'\x}] - \z^{\x}_q\|_2^2 + \lambda_{\text{commit}} \|\z^{'\x} - \text{sg}[\z^{\x}_q]\|_2^2}\\
    &\hspace{-13mm}:= \Loss_{Wl_2}  + \Loss_{\text{perc}} + \lambda_{\text{GAN}} \Loss_{\text{GAN}} + \lambda_{\text{vq}} \Loss_{\text{vq}} + \lambda_{\text{commit}} \Loss_{\text{commit}}   \label{Total image loss}
\end{align}
\FP{where the last two terms and the $\lambda$'s are defined as in \eqref{Total semantic loss}, while the first three terms are defined as follows.} 
The weighted $\ell_2$ loss $\Loss_{Wl_2}$, defined as:
\begin{equation} 
\Loss_{\text{W}l_2}(\x, \hat{\x}) = \frac{1}{H W}\sum_{(h,w)} \text{w}_{\s_{(h,w)}} \| \x_{(h,w)} - \hat{\x}_{(h,w)} \|^2, \label{eq: weighted_l2_loss} 
\end{equation}
is designed to adjust the importance of different semantic classes in the image. 
In \eqref{eq: weighted_l2_loss}, $\x_{(h,w)}$ and $\hat{\x}_{(h,w)}$ are the pixel values at location $(h,w)$ of the tensors $\x$ and $\hat{\x}$, respectively, and $\text{w}_{\s_{(h,w)}}$ is the weight associated with the semantic class at that pixel, as given by the \gls{ssm} $\s$. 
Specifically, while training the model used in our numerical examples, we used the following weights: 

\vspace{0.2cm}
\begin{tabular}{ll}
    \textbullet \;\; $\text{w} = 1$    & for the relevant classes "traffic signs" \\ &  and "traffic lights". \\
    \textbullet \;\; $\text{w} = 0.55$ & for the classes "people" and "rider". \\
    \textbullet \;\; $\text{w} = 0$    & for the non-relevant classes "sky"  \\ &  and "vegetation". \\
    \textbullet \;\; $\text{w} = 0.15$ & for all other classes.
\end{tabular}\\

Of particular interest is the zero weight assigned to the classes "sky" and "vegetation". This allows the network to neglect 
the pixel-by-pixel reconstruction of these classes. In fact, the real colors of the sky and trees for the task at hand are semantically irrelevant. 

The perceptual loss $\Loss_{\text{perc}}$ is instead the part that ensures that the reconstructed image maintains visual similarity to the original, preventing unrealistic alterations such as an unnatural sky color. To quantify this loss, we used  \gls{lpips} \cite{Zhang2018LPIPS}, which evaluates the difference between two images in a latent space representation. The underlying idea is that if two images convey similar semantic content, for example, if they both contain relevant objects such as pedestrians, cars, or traffic lights, each with the correct shape and positioned appropriately relative to one another, the semantic distortion should be low, regardless of any differences at the pixel level. LPIPS, instead of computing pixel-wise differences between $\x$ and $\hat{\x}$,   
compares the latent features extracted from the images by employing a deep neural network (DNN) pre-trained for object recognition. The DNN
is somehow substituting a human to quantify the perceptual difference between two images. 
Formally, LPIPS is defined as:
\begin{equation}
\label{LPIPS}
    \Loss_{\text{perc}} = \sum_{l\in L} \frac{1}{H_l W_l} \sum_{(h_j,w_j)} \left\| \text{w}_l \odot  \left( \phi_l(\x) - \phi_l(\hat{\x})  \right) \right\|_2^2
\end{equation}
where $\phi_l(\cdot)$ denotes the activation from the $l$-th layer of a pre-trained network $\phi$, $L$ is the set of layers used for feature extraction, $H_l$ and $W_l$ are the height and width of the feature map at layer $l$, and $\text{w}_l$ are learned weights that adjust the contribution of each layer. The operator $\odot$ denotes element-wise multiplication between the weight $\text{w}_l$ and all the elements of the difference $\left( \phi_l(\x) - \phi_l(\hat{\x})  \right)$ in any location coordinate $(h_j,w_j)$. By comparing the feature representations at multiple layers, \glspl{lpips} captures perceptual differences at different scales and abstraction levels. Lower \gls{lpips} values indicate better reconstruction quality. 

Another important novel aspect of the proposed approach is the 
adversarial loss $\mathcal{L}_{\text{GAN}}$ involving a new semantic-aware discriminator network. 
The discriminator $D_{disc}^\x$ plays a crucial role in adversarial training by determining whether an image reconstructed by $G_\x$ is real or fake.  However, a potential drawback is that the discriminator might focus on non-relevant parts of the image to produce its classification score. 
For instance, it might prioritize vegetation details, and classify images as real only if the leaves on the trees have a certain level of detail. This will force the generator $G_\x$ to reconstruct images with better vegetation details to fool the discriminator. Unfortunately, this is not optimal for the structure of the \gls{sqgan}. Giving more importance to the vegetation will decrease the importance of other classes, thus causing the \gls{samm} module  to select the wrong latent vectors as relevant.

In recent years, several works have proposed ways to modify the discriminator by introducing 
various conditioning. For example, \cite{Oluwasanmi2020condDiscr} conditioned the discriminator on the \gls{ssm} to improve \gls{sseg} retention, while \cite{Chen2020ssd} enforced the discriminator to focus on high frequency components. However, these methods do not adequately address the issue at hand. 
To achieve the desired performance, it is essential to adjust the discriminator to minimize its focus on non-relevant regions. To this end, we make the following observations about the discriminator's behavior:
\begin{itemize}
    \item For a (generic) trained discriminator $D_{disc}$ and two images $\x$ and $\y$, $D_{disc}(\x)$ is likely similar to $D_{disc}(\y)$ if both images originate from the same data distribution, i.e., $p_{\x} = p_{\y}$. However, $D_{disc}(\x) = D_{disc}(\y)$ is not guaranteed unless the images are identical or indistinguishable by the discriminator.
    \item If $p_{\y}$ differs from $p_{\x}$, $D_{disc}(\y)$ will likely differ from $D_{disc}(\x)$. The output difference is primarily influenced by the aspects of $p_{\y}$ that deviate from $p_{\x}$.
    \item The greater the difference between $p_{\x}$ and $p_{\y}$, the higher the uncertainty in predicting $D_{disc}(\y)$ based on $D_{disc}(\x)$.
\end{itemize}
Based on these insights, we propose to reduce the discriminator's focus on non-relevant semantic classes.
Our approach consists of artificially modifying the reconstructed image $\hat{\x}$ before it is evaluated by the discriminator \FP{and in \eqref{Total image loss} has been represented as conditioning the input of the discriminator as $D_{disc}^\x(\cdot|\s)$}. This modification aims to minimize the differences in non-relevant regions between $\x$ and $\hat{\x}$, bringing the data distribution of these regions of $\hat{\x}$ closer to the real distribution of $\x$. For example, if the generator reconstructs a tree with dark green leaves when in reality they are light green, the color in $\hat{\x}$ will be artificially shifted toward light green. 

This artificial editing is performed by considering the residual between the real and reconstructed images, defined as $\br = \x - \hat{\x}$. \FP{By using the \gls{ssm} to mask the} pixel-wise difference between the two images, and adding back a fraction of the residual to $\hat{\x}$ it is possible to obtain the new image $\hat{\x}_{rel} = \hat{\x} + \w_{rel} \odot \br$. \FP{In this context $\w_{rel}$ represents the re-scaling relevance tensor that is obtained as a function of the \gls{ssm} $\s$ and has the same shape as for the images, namely shape $3 \times H \times W$. The element $(i,h,w)$ of the tensor $\w_{rel}$ is obtained from $\s$ by assigning the following values based on the class of the pixel $(h,w)$. In particular, in the training for our numerical examples, we used:}

\vspace{0.2cm}
\begin{tabular}{ll}
    \textbullet \;\; $\text{w} = 0.90$    & for the class "sky". \\
    \textbullet \;\; $\text{w} = 0.80$ & for the class "vegetation". \\
    \textbullet \;\; $\text{w} = 0.40$    & for the class "street". \\
    \textbullet \;\; $\text{w} = 0$ & for the other classes.
\end{tabular}\\

For example, this means that  80\% of the difference between the shades of green leaves is removed before presenting the image to the discriminator. The modified vegetation in $\hat{\x}_{rel}$ will appear much closer to the real light green in $\x$, thus reducing the discriminator's focus on this non-relevant region.
It is important to acknowledge that this approach negatively impacts the generator's ability to accurately reproduce these specific non-relevant classes. Nevertheless, the perceptual loss still considers the entire image, guiding the generator to reconstruct the "sky," "vegetation," and "streets" to maintain overall realism.

The sub-network $G_\x$ is trained for different masking fractions $m_\x$, \FP{ similarly to $G_\s$, to increase the robustness of the model with respect to different conditions of the channel}. These values are selected from a finite set ranging from $5\%$ to $100\%$ with an expected value of $35\%$.
The structure of $D_{disc}^\x$ is the same as $D_{disc}^\s$ and the training is performed using the Adam optimizer with a learning rate of $10^{-4}$ with batch size of $8$. The model is trained for $200$ epochs with early stopping.

\subsection{Fine-tuning $G$}\label{sec: SQGAN training G}

The final step involves fine-tuning the entire network $G$ by freezing the parameters of $G_\s$ and updating only those of $G_\x$. This fine-tuning addresses the scenario where the original \gls{ssm} $\s$ is unavailable at the receiver, that is, when the \gls{sqgan} is used for \gls{sc}.
Notice that the fine-tuning step is {\em essential} since the quality of $\hat{\s}$ is influenced by the masking fraction $m_\s$, and this dependency is missed in the separate training steps described above.
The loss function and the semantic-aware discriminator $D_{disc}^\x$ remain identical to those used in the training of $G_\x$. Fine-tuning is conducted over 100 epochs with early stopping to prevent overfitting. During each iteration, both masking fractions $m_\x$ and $m_\s$ are randomly selected from the same distribution. The Adam optimizer is employed with a learning rate of $10^{-4}$ and a batch size of 8.

\section{Results} \label{sec: SQGAN numerical results}

This section focuses on the performance of the proposed \gls{sqgan} evaluated using 500 images (and corresponding \glspl{ssm}) 
from the validation set of the Cityscapes dataset \cite{Cordts2016Cityscapes}.
In \sref{sec: SQGAN result samm},  \textcolor{black}{we illustrate the function and role of the SAMM in selecting features of $\x$ and $\s$ that are semantically relevant. Furthermore, we also show the effect of masking fractions on the final reconstruction.} In \sref{sec: SQGAN result comparison} the performance of the proposed \gls{sqgan} is compared with \gls{sota} compression algorithms like \gls{bpg}, \gls{jpeg2000} \FP{and deep-learning based alternatives such as \gls{hific} \cite{mentzer2020hific} and \gls{fcc} \cite{Iwai2021fcc}}. \FP{For the selection of the deep learning-based comparison methods, we considered only those  providing the open source code and the model weights. Furthermore, we have not considered methods designed for JSCC, since 
we consider only source coding.} 

\FP{We also computed the \gls{ssm} $\tilde{\s}$ of the reconstructed image $\hat{\x}$ and compared it with the \gls{ssm} $\s$ of the original image, to quantitatively assess the capability of the coding method to preserve some semantics aspects of the image to be transmitted, such as
identification of objects like "pedestrians", "traffic signs", etc., the estimation of their shape and of their relative positioning. 
}
\subsection{SAMM function}\label{sec: SQGAN result samm}
\begin{figure*}[!t]
    \centering
    \adjustbox{valign=t}{\includegraphics[width=0.45\textwidth]{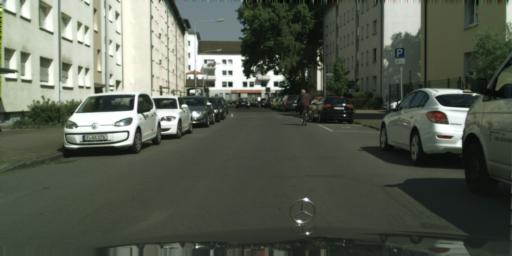}}%
    \hspace{5mm}
    \adjustbox{valign=t}{\includegraphics[width=0.45\textwidth]{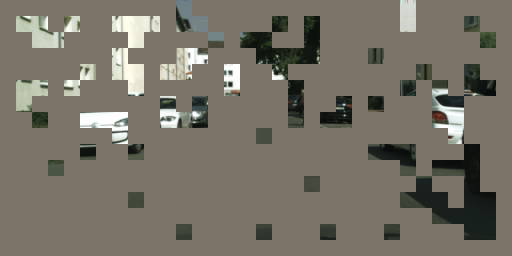}}\\[1mm]
    \adjustbox{valign=t}{\includegraphics[width=0.45\textwidth]{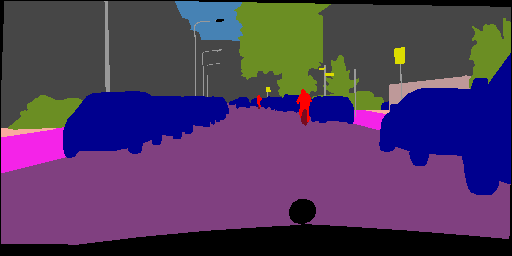}}%
    \hspace{5mm}
    \adjustbox{valign=t}{\includegraphics[width=0.45\textwidth]{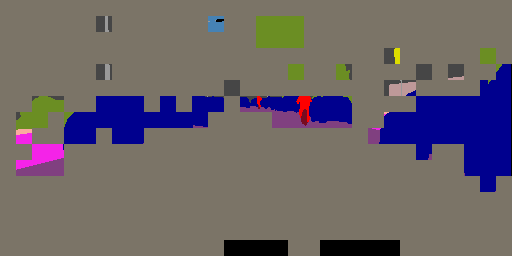}}
    \caption[Visual representation of the effect of the \acrshort{samm} module]{Visual representation of the latent tensor selection of the \acrshort{samm} projected in the image and \acrshort{ssm} domain. In both cases the masking has been fixed to $m_\x=m_\s=0.20$ and the regions considered semantically relevant are shown on the right.}
    \label{fig: SQGAN masked images}
\end{figure*}

\gls{sqgan} adopts various techniques to force the model to correctly reconstruct the relevant regions of the image. The weighted loss function based on the semantic classes, the semantic-aware discriminator $D_{disc}^\x$, and the data augmentation step described in \ref{sec: SQGAN Data Augmentation}, have been designed with this idea in mind. Their scope is to guide \gls{samm} to identify and select the relevant latent vectors $\z_k^\x$ and $\z_k^\s$. Since the selection is performed on the latent tensor, \textcolor{black}{to understand the effect of SAMM, we project back from the latent tensor to the image (pixel) domain \cite{Huang2023MaskedVQ-VAE}. \fref{fig: SQGAN masked images} shows such a projection for a specific example.} \footnote{The projection shown in figure \ref{fig: SQGAN masked images} was implemented by leveraging the convolutional structure of the two encoders that allows to trace back the spatial correlation between the latent tensor and the original frame.} 
The left column displays the original $\x$ and $\s$, whereas the right column indicates the regions associated with the latent vectors deemed most relevant by \gls{samm}. In this example, the masking fractions are chosen as $m_\x=m_\s=0.20$.

It is immediately evident that $G_\x$ and $G_\s$ consider different regions as relevant. The \gls{samm} in $G_\s$ focuses on regions with the most change in semantic classes. Streets, buildings, and the sky require very few associated latent vectors for their representation. In contrast, the regions of the \gls{ssm} containing semantically relevant classes are strongly prioritized. 

Similarly, the \gls{samm} in $G_\x$ shows a strong preference for relevant classes like cars and people. However, it also focuses on areas previously ignored, such as the street and the buildings. Indeed, thanks to the conditioning on the \gls{ssm}, the sub-network \( G_\x \) is aware of the location and shape of each relevant object, allowing it to focus on other aspects such as colors and textures. As a result, elements like the streets and the sky still require some latent vectors to accurately reconstruct their colors.
 However, most of the latent vectors are selected from the truly relevant regions.

In both cases, \gls{samm} tends to prefer regions that contain more semantically relevant objects. This is a direct effect of the various techniques adopted to train the model as described in \sref{sec: SQGAN training}. \\

\textcolor{black}{Next, we examine the effect of different masking fractions on the quality of the reconstructed image and SSM.} In fact increasing $m_\x$ and $m_\s$ is expected to increase the overall quality of $\hat{\x}$ and $\hat{\s}$ respectively, while decreasing is expected to do the opposite. \fref{fig: SQGAN visual result changing masking} shows visually how masking fractions influence the reconstructed outputs.
\begin{figure*}[!t]
    \centering
    \begin{tabular}{>{\centering\arraybackslash}m{0.24\textwidth} 
                    @{\hspace{1mm}}>{\centering\arraybackslash}m{0.24\textwidth} 
                    @{\hspace{1mm}}>{\centering\arraybackslash}m{0.24\textwidth} 
                    @{\hspace{1mm}}>{\centering\arraybackslash}m{0.24\textwidth}}
        Original & 
        $m_\x=0.95, \;m_\s=0.15$ & 
        $m_\x=0.55, \;m_\s=0.55$ & 
        $m_\x=0.15, \;m_\s=0.95$ \\
    \end{tabular}
    \\

    \begin{tabular}{>{\centering\arraybackslash}m{0.24\textwidth} 
                    @{\hspace{1mm}}>{\centering\arraybackslash}m{0.24\textwidth} 
                    @{\hspace{1mm}}>{\centering\arraybackslash}m{0.24\textwidth} 
                    @{\hspace{1mm}}>{\centering\arraybackslash}m{0.24\textwidth}}
        \begin{tikzpicture}
            \node[anchor=south west,inner sep=0] (image1) at (0,0) {\includegraphics[width=\linewidth]{Figures/Semantic_MQ-GAN/real_image.png}};
            \node[anchor=south east,inner sep=0] at (image1.south east) {\includegraphics[width=0.45\linewidth]{Figures/Semantic_MQ-GAN/real_ssm.png}};
        \end{tikzpicture} & 
        \begin{tikzpicture}
            \node[anchor=south west,inner sep=0] (image4) at (0,0) {\includegraphics[width=\linewidth]{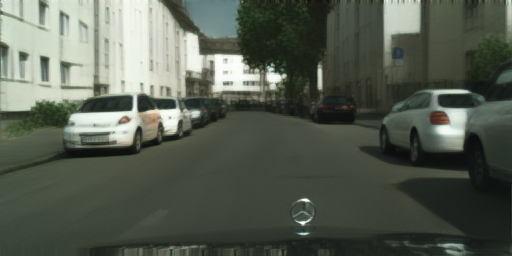}};
            \node[anchor=south east,inner sep=0] at (image4.south east) {\includegraphics[width=0.45\linewidth]{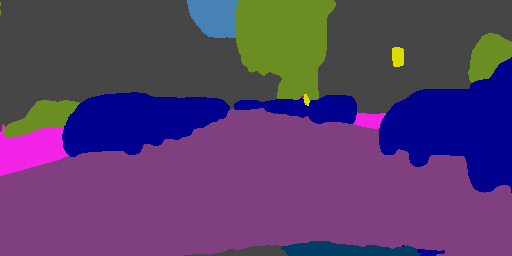}};
        \end{tikzpicture} & 
        \begin{tikzpicture}
            \node[anchor=south west,inner sep=0] (image3) at (0,0) {\includegraphics[width=\linewidth]{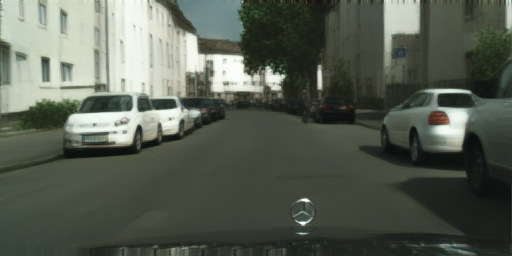}};
            \node[anchor=south east,inner sep=0] at (image3.south east) {\includegraphics[width=0.45\linewidth]{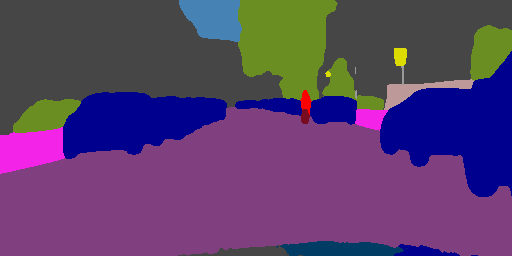}};
        \end{tikzpicture} & 
        \begin{tikzpicture}
            \node[anchor=south west,inner sep=0] (image2) at (0,0) {\includegraphics[width=\linewidth]{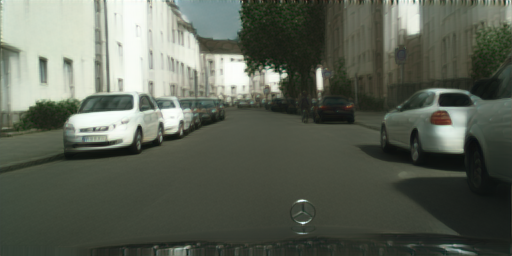}};
            \node[anchor=south east,inner sep=0] at (image2.south east) {\includegraphics[width=0.45\linewidth]{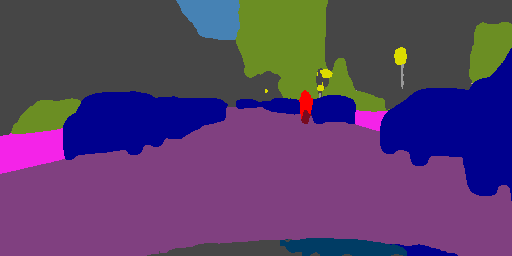}};
        \end{tikzpicture} \\

        \includegraphics[width=\linewidth]{Figures/Semantic_MQ-GAN/real_ssm.png} & 
        \includegraphics[width=\linewidth]{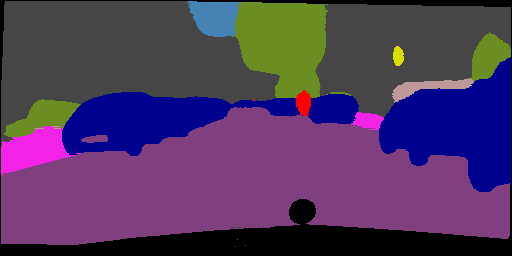} & 
        \includegraphics[width=\linewidth]{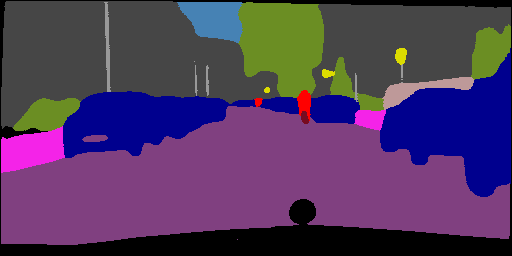} & 
        \includegraphics[width=\linewidth]{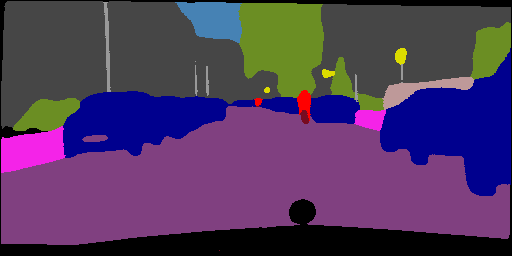} \\
    \end{tabular}
    
    \caption[Visual comparison between different results at different masking fractions]{Visual comparison between the same image and \acrshort{ssm} at different masking reactions $m_\x$ and $m_\s$. The original image and \acrshort{ssm} are shown on the left. The upper row shows the reconstructed $\hat{\x}$ and the generated \gls{ssm} using the \gls{sota} \gls{ssmodel} INTERN-2.5 \cite{Wang2022internimage}. The bottom row shows the reconstructed \gls{ssm} $\hat{\s}$. All pairs $(\hat{\x}, \hat{\s})$ are obtained at $0.05$\gls{bpp}.}
    \label{fig: SQGAN visual result changing masking}
\end{figure*}

The column on the left shows the original $\x$ and $\s$. The other columns show on top the reconstructed $\hat{\x}$ and the generated \gls{ssm} obtained from $\hat{\x}$ via the INTERN-2.5 \gls{ssmodel} \cite{Wang2022internimage}. The bottom row shows the reconstructed $\hat{\s}$. 
These examples are obtained by fixing the compression level to a total amount of $0.05$ \gls{bpp} and by letting the masking fractions $m_\x$ and $m_\s$ to vary, while keeping their sum constant. In this example, $m_\x + m_\s = 1.1$.\begin{figure*}[!h]
    \centering
    \begin{minipage}[b]{0.45\textwidth}
        \centering
        \includegraphics[width=\textwidth]{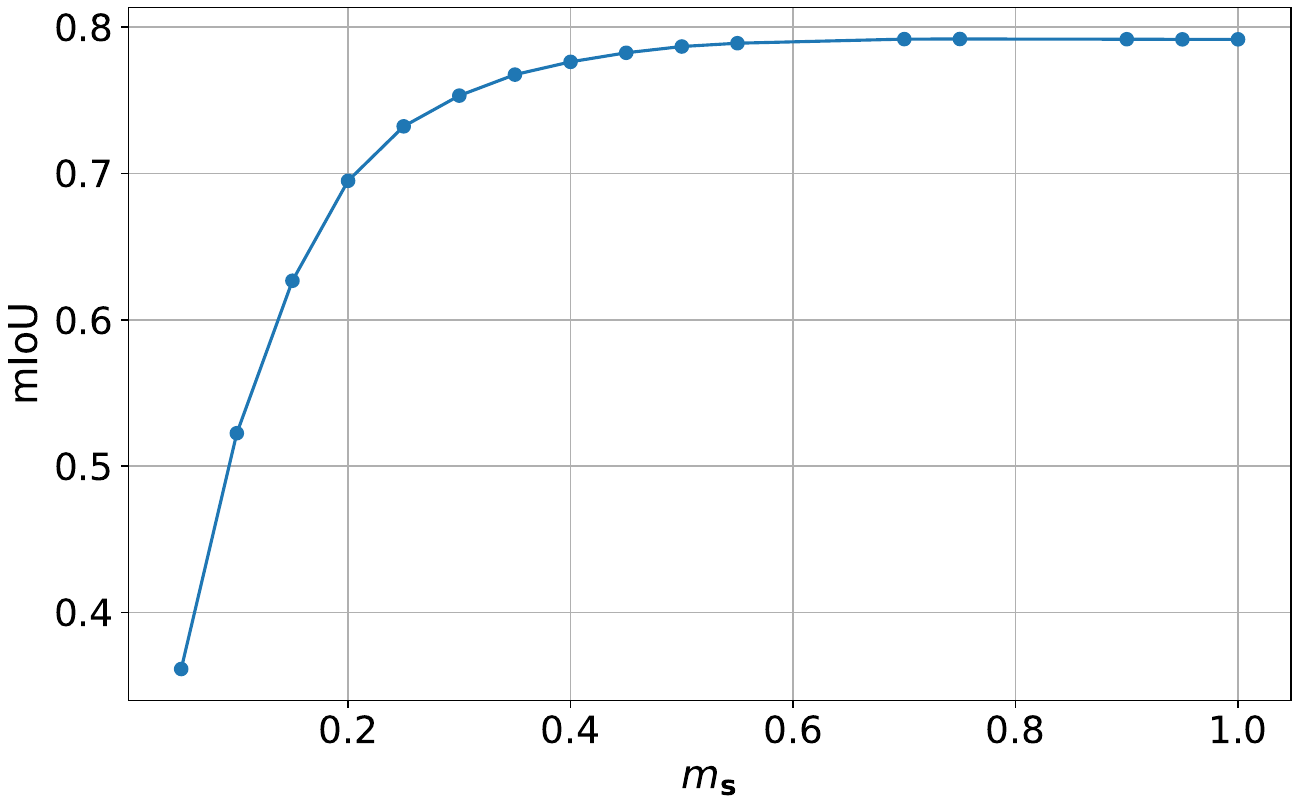}
        \caption[\acrshort{ssm} retention as a function of the masking fraction]{\acrshort{ssm} retention  evaluated as the mIoU between the true $\s$ and the reconstructed $\hat{\s}$ as a function of the masking fraction $m_\s$. As the masking fraction $m_\s$ increases the network $G_\s$ is able to better reconstruct the \gls{ssm}. However, the increase of performance reaches a plateau from $m_\s \geq 0.2$ ($BPP_\s=0.011$\gls{bpp}).}
        \label{fig: SQGAN miou vs m_s}
    \end{minipage}
    \hfill
    \begin{minipage}[b]{0.45\textwidth}
        \centering
        \includegraphics[width=\textwidth]{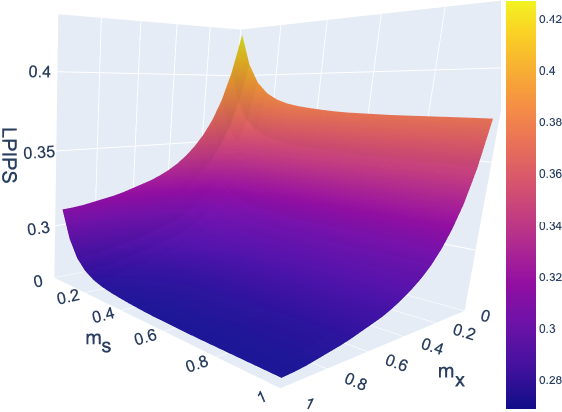}
        \caption[\acrshort{lpips} as a function of the masking fractions]{\acrshort{lpips} evaluated between $\x$ and $\hat{\x}$ as the masking fractions $m_\s$ and $m_\x$ vary. By properly selecting the  masking fractions it is possible to reduce the redundancy, while preserving a task-specific feature, represented by the LPIPS value }
        \label{fig: SQGAN lpips 3d plot}
    \end{minipage}
\end{figure*}

The quality of $\hat{\x}$ is significantly affected by the fidelity of $\hat{\s}$, which depends heavily on $m_\s$, more than $m_\x$. A reduction in object detail within $\hat{\s}$ directly limits the ability of $\hat{\x}$ to retain such details. This is evident by comparing the bottom row $\hat{\s}$ with the generated \gls{ssm} in the upper row: any detail absent in $\hat{\s}$ is also absent in the \gls{ssm}. \\
Conversely, the influence of $m_\x$ on the overall quality exhibits the expected behavior: increasing $m_\x$ enhances image fidelity, particularly for non-relevant details. Relevant features are prioritized and reconstructed effectively even at low values of $m_\x$, whereas finer details, such as building windows, are better preserved when $m_\x=0.95$. This demonstrates the model's capability to prioritize semantically relevant features before refining non-relevant aspects of the image.
\\

To assess the quality of the reconstructed SSM, we use the mean Intersection over Union (mIoU) parameter, defined as:
\begin{equation} 
\text{mIoU} = \frac{1}{n_c} \sum_{i=1}^{n_c} \frac{|\s_i \cap \s'_i|}{|\s_i \cup \s'_i|}, 
\end{equation}
where $n_c$ represents the number of semantic classes, and $\s_i$ and $\s'_i$ are pixel sets of class $i$ in $\s$ and the predicted $\s'$, respectively. This parameter measures how well two segmentation maps overlap with each other. A value close to $1$ indicates a high semantic content retention in $\hat{\x}$. 
\begin{figure*}[!h]
    \centering
    \begin{tabular}{>{\centering\arraybackslash}m{0.24\textwidth} 
                    @{\hspace{1mm}}>{\centering\arraybackslash}m{0.24\textwidth} 
                    @{\hspace{1mm}}>{\centering\arraybackslash}m{0.24\textwidth} 
                    @{\hspace{1mm}}>{\centering\arraybackslash}m{0.24\textwidth}}
        Original & 
        $0.019$\textit{BPP} & 
        $0.038$\textit{BPP} & 
        $0.078$\textit{BPP} \\
    \end{tabular}
    \\

    \begin{tabular}{>{\centering\arraybackslash}m{0.24\textwidth} 
                    @{\hspace{1mm}}>{\centering\arraybackslash}m{0.24\textwidth} 
                    @{\hspace{1mm}}>{\centering\arraybackslash}m{0.24\textwidth} 
                    @{\hspace{1mm}}>{\centering\arraybackslash}m{0.24\textwidth}}
        \begin{tikzpicture}
            \node[anchor=south west,inner sep=0] (image1) at (0,0) {\includegraphics[width=\linewidth]{Figures/Semantic_MQ-GAN/real_image.png}};
            \node[anchor=south east,inner sep=0] at (image1.south east) {\includegraphics[width=0.45\linewidth]{Figures/Semantic_MQ-GAN/real_ssm.png}};
        \end{tikzpicture} & 
        \begin{tikzpicture}
            \node[anchor=south west,inner sep=0] (image2) at (0,0) {\includegraphics[width=\linewidth]{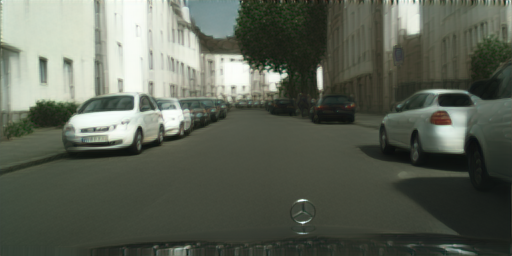}};
            \node[anchor=south east,inner sep=0] at (image2.south east) {\includegraphics[width=0.45\linewidth]{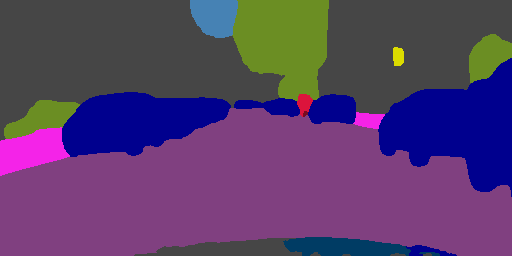}};
        \end{tikzpicture} & 
        \begin{tikzpicture}
            \node[anchor=south west,inner sep=0] (image3) at (0,0) {\includegraphics[width=\linewidth]{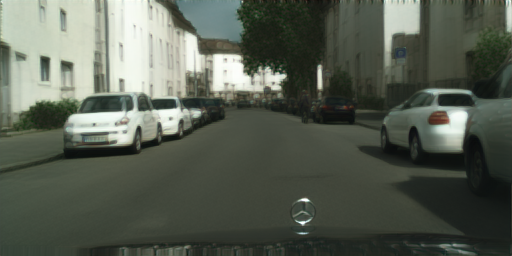}};
            \node[anchor=south east,inner sep=0] at (image3.south east) {\includegraphics[width=0.45\linewidth]{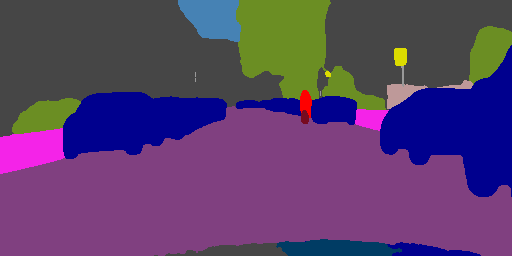}};
        \end{tikzpicture} & 
        \begin{tikzpicture}
            \node[anchor=south west,inner sep=0] (image4) at (0,0) {\includegraphics[width=\linewidth]{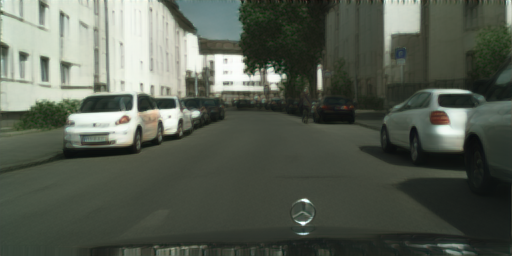}};
            \node[anchor=south east,inner sep=0] at (image4.south east) {\includegraphics[width=0.45\linewidth]{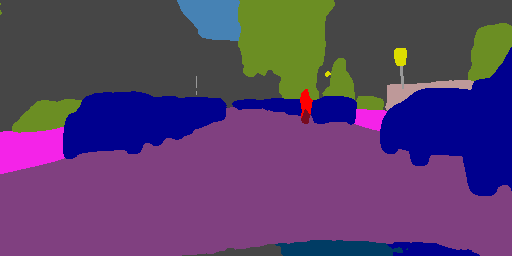}};
        \end{tikzpicture} \\
        \begin{tikzpicture}
            \node[anchor=south west,inner sep=0] (image5) at (0,0) {\includegraphics[width=\linewidth]{Figures/Semantic_MQ-GAN/real_image.png}};
            \node[anchor=south east,inner sep=0] at (image5.south east) {\includegraphics[width=0.45\linewidth]{Figures/Semantic_MQ-GAN/real_ssm.png}};
        \end{tikzpicture} & 
        & 
        \begin{tikzpicture}
            \node[anchor=south west,inner sep=0] (image7) at (0,0) {\includegraphics[width=\linewidth]{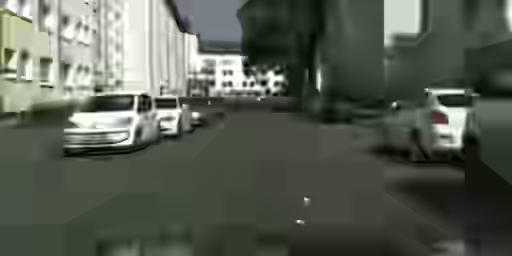}};
            \node[anchor=south east,inner sep=0] at (image7.south east) {\includegraphics[width=0.45\linewidth]{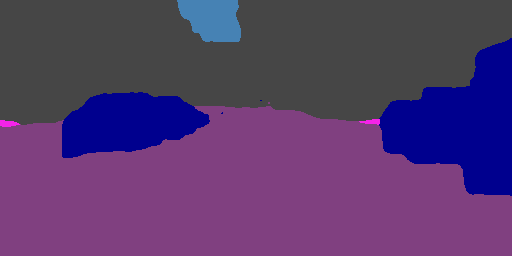}};
        \end{tikzpicture} & 
        \begin{tikzpicture}
            \node[anchor=south west,inner sep=0] (image8) at (0,0) {\includegraphics[width=\linewidth]{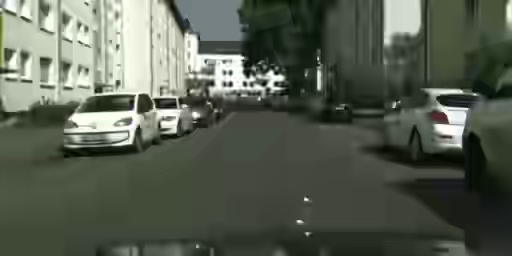}};
            \node[anchor=south east,inner sep=0] at (image8.south east) {\includegraphics[width=0.45\linewidth]{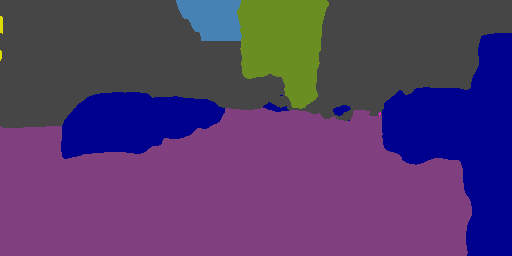}};
        \end{tikzpicture} \\
    \end{tabular}
    
    \caption[Visual comparison between \acrshort{bpg} and \acrshort{sqgan}]{Visual comparison at different compression rates between the proposed \acrshort{sqgan} (TOP) and the classical \acrshort{bpg} (BOTTOM). The \glspl{ssm} shown are generated from $\hat{\x}$ via the \acrshort{sota} \acrshort{ssmodel} INTERN-2.5 \cite{Wang2022internimage}. The proposed model is able to reconstruct images with higher semantic retention and lower values of \acrshort{bpp} compared with \acrshort{bpg}. The \acrshort{bpg} algorithm is not able to compress images at lower values than $0.038$ \acrshort{bpp}, thus the comparison is limited to $0.038$ and $0.078$ \acrshort{bpp}.}
    \label{fig: SQGAN visual comparison sqgan bpg}
\end{figure*}

The impact of $m_\s$ on the mIoU is illustrated 
in \fref{fig: SQGAN miou vs m_s}. 
It is useful to notice that, for $m_\s \geq 0.55$, the value of mIoU remains almost constant. This means that, above that value, additional latent vectors do not add relevant semantic features.
For very low values of $m_\s$ the performance is poor, but, as soon as $m_\s \geq 0.20$, the model is able to reconstruct $\hat{\s}$ with an acceptable level of semantic retention. To give a term of comparison, the value of $m_\s =0.20$  corresponds to a compression of $BPP_\s=0.011$ \gls{bpp}. This means that at this low value of \gls{bpp} the model is already able to preserve valuable details in the \gls{ssm}. Increasing $m_\s$ beyond the value of 0.55 (equivalent to $BPP_\s=0.025$ \gls{bpp}) does not provide further improvement. We interpret this saturation as follows: beyond this value, all semantically relevant latent vectors have been selected and quantized. Adding other vectors will only increase the amount of redundant information, without improving the reconstructed \gls{ssm}. This behavior is quite different from what was observed and discussed previously about $m_\x$, where even for values close to $1$, the improvements were still visible on the final output.

It is also interesting to see how both $m_\x$ and $m_\s$ affect $\hat{\x}$. For this purpose, in \fref{fig: SQGAN lpips 3d plot} we report the value of \gls{lpips} to measure the distance between $\x$ and $\hat{\x}$ \FP{and show the importance of semantic masking}. As expected, the performance along the $m_\s$ axis has a strong influence on the output only for values of $m_\s \leq 0.55$. Instead, the influence of $m_\x$ is observed along the whole range from 0 to 1. This is consistent with the visual results shown in \fref{fig: SQGAN visual result changing masking} \FP{and show how, by properly selecting the masking coefficients $m_x$ and $m_s$, we can strongly reduce the redundancy, while preserving a task-specific feature, represented by the LPIPS value.}

\subsection{Visual Results and Comparisons with SOTA image compression}\label{sec: SQGAN result comparison}
In this section, the results of the proposed \gls{sqgan} are compared with the SOTA \FP{conventional algorithms \gls{bpg} and \gls{jpeg2000}, and with deep-learning-based compression algorithms  \gls{hific} \cite{mentzer2020hific} and \gls{fcc} \cite{Iwai2021fcc}}. A visual comparison between \gls{bpg} and \gls{sqgan} is shown in \fref{fig: SQGAN visual comparison sqgan bpg}. 

The top row represents the reconstructed image $\hat{\x}$ obtained with the proposed \gls{sqgan} and the associated \gls{ssm} generated via the INTERN-2.5 \gls{ssmodel}. The bottom row shows the reconstructed image obtained by using the \gls{bpg} algorithm and the associated generated \gls{ssm}. 
\begin{figure*}[!t]
    \centering
    \begin{subfigure}[t]{0.45\textwidth}
        \centering
        \includegraphics[width=\textwidth]{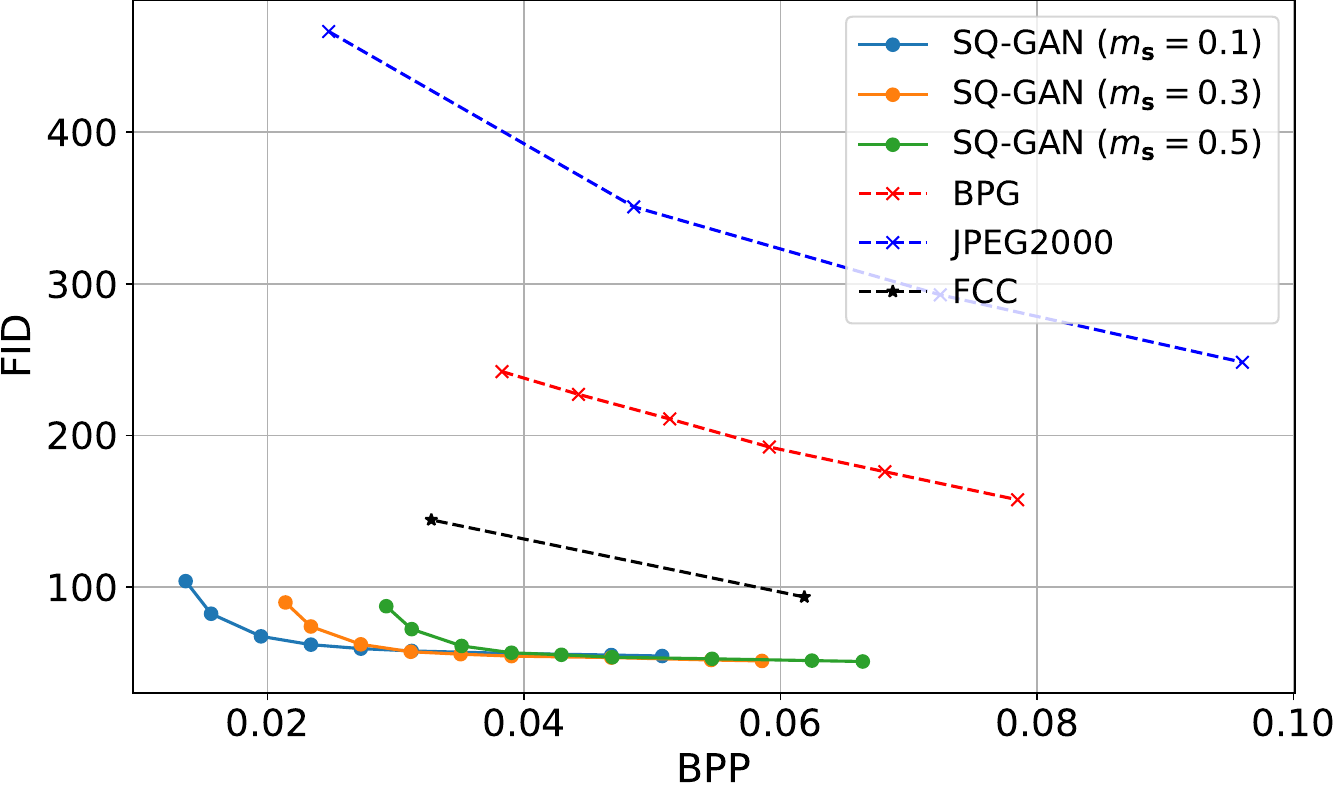}
        \caption*{(a)} 
    \end{subfigure}%
    \hspace{5mm}
    \begin{subfigure}[t]{0.45\textwidth}
        \centering
        \includegraphics[width=\textwidth]{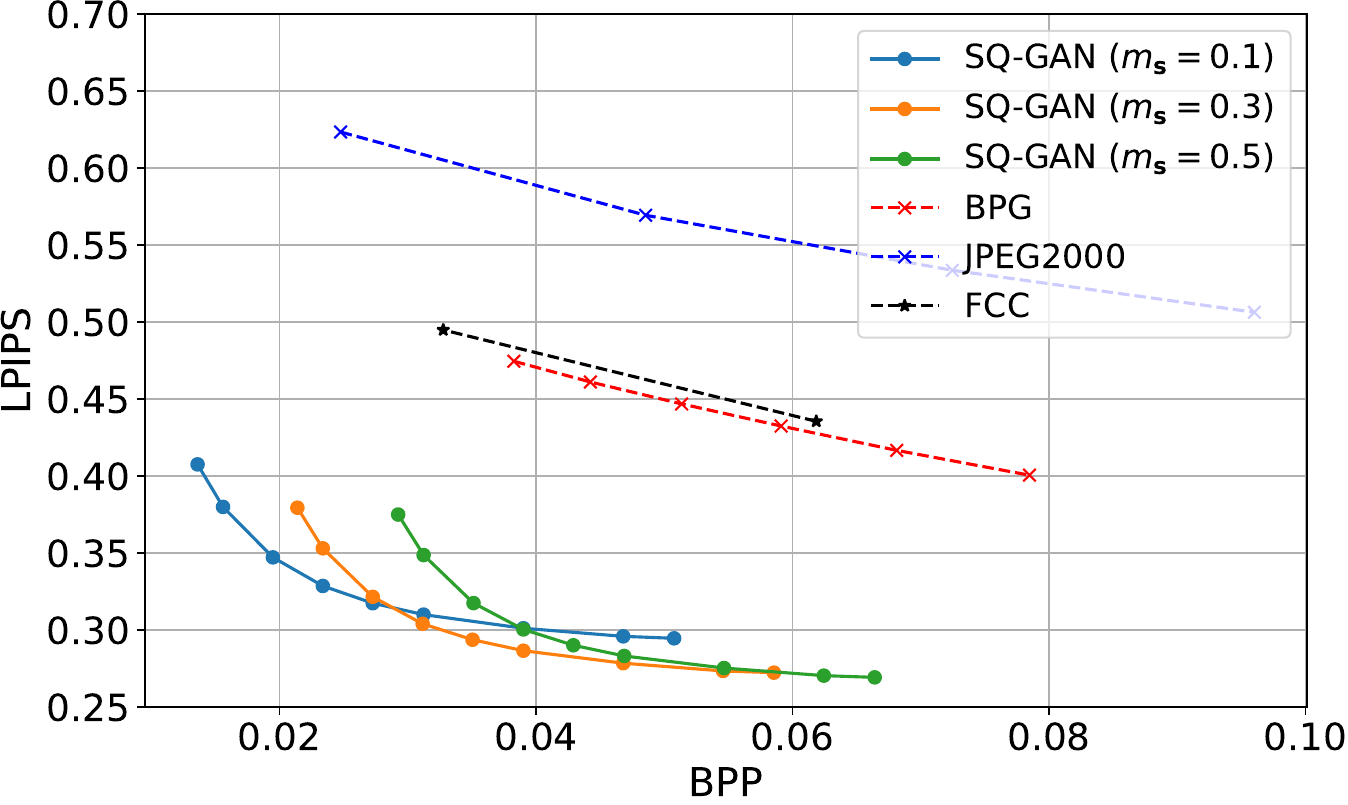}
        \caption*{(b)}
    \end{subfigure}
    
    \vspace{0.05mm} 

    \begin{subfigure}[t]{0.45\textwidth}
        \centering
        \includegraphics[width=\textwidth]{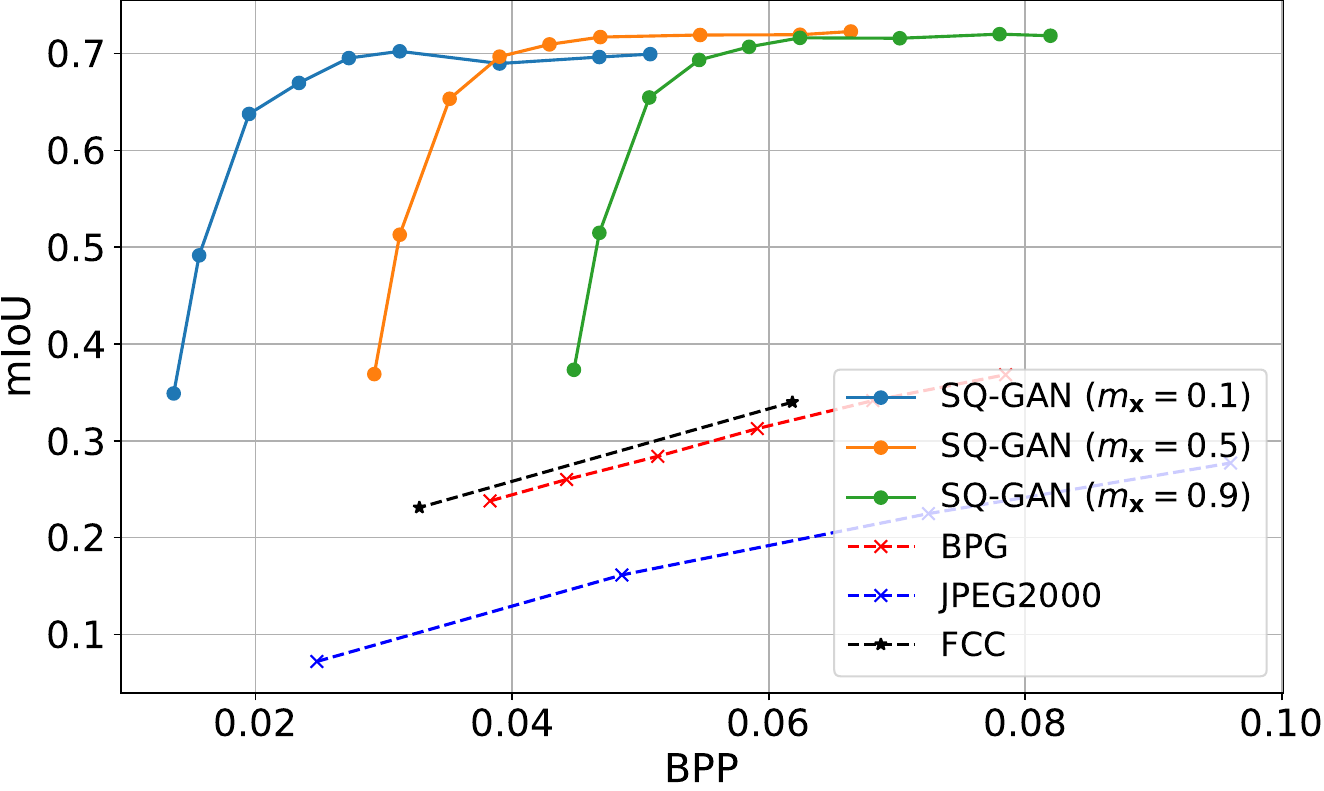}
        \caption*{(c)}
    \end{subfigure}%
    \hspace{5mm}
    \begin{subfigure}[t]{0.45\textwidth}
        \centering
        \includegraphics[width=\textwidth]{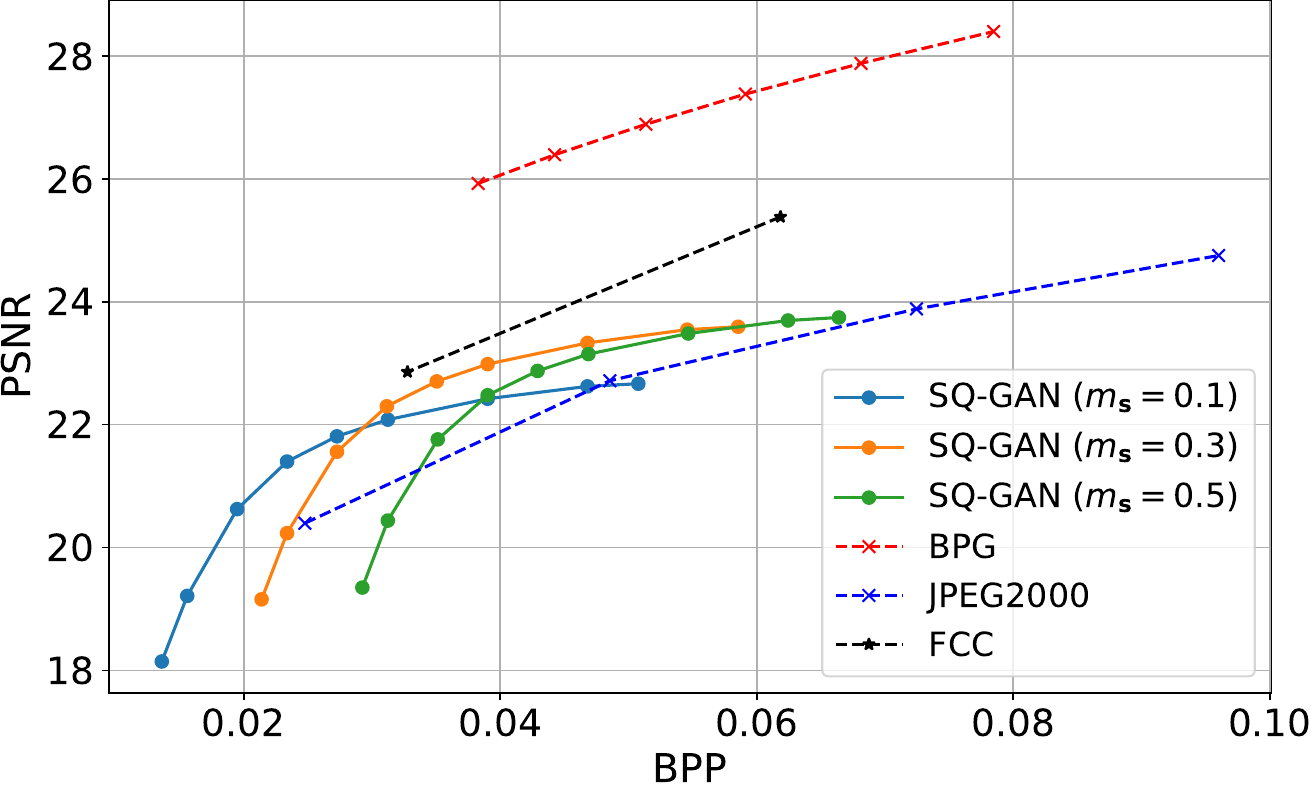}
        \caption*{(d)}
        \label{fig: SQGAN psnr vs bpp}
    \end{subfigure}

    \caption[Performance comparison between the \acrshort{sqgan} and classical compression algorithms]{Performance comparison between \acrshort{bpg}, \acrshort{jpeg2000} and \acrshort{sqgan} in terms of semantic metrics and the classic pixel-by-pixel \acrshort{psnr}.}
    \label{fig: SQGAN all metrics comparison}
\end{figure*}
We notice that \gls{sqgan} can reach lower values of \gls{bpp}. In particular, in this case BPG cannot compress images at \gls{bpp} lower than $0.038$. Furthermore, while \gls{bpg} uses precious resources to reconstruct semantically irrelevant details such as the windows of the buildings,  \gls{sqgan} focuses on the relevant parts. This is further evidenced by the amount of semantic retention of the generated \gls{ssm}. For example, the cyclist on the bike is still visible and can be correctly identified by \gls{ssmodel}, at all levels of BPP in \fref{fig: SQGAN visual comparison sqgan bpg}. In contrast, the cyclist completely disappears in the generated SSM from the BPG reconstructed image at all levels of BPP in \fref{fig: SQGAN visual comparison sqgan bpg}.\\
As a comparison term, to obtain a level of semantic retention similar to the one obtained by \gls{sqgan} at $0.038$ \gls{bpp}, \gls{bpg} requires a rate of $0.280$ \gls{bpp}.\\

As a further parameter used to quantify the (semantic) distortion between the original and reconstructed images, we used the Fréchet Inception Distance (FID), evaluated over a batch of images  \cite{Heusel2017FID}.
The value of FID
is defined as \cite{Heusel2017FID}: 
\begin{align}
    \text{FID} &= \left\| \mu_{\phi}(\mathbf{X}) - \mu_{\phi}(\hat{\mathbf{X}}) \right\|_2^2 \\
    &+ \text{Tr}\left( \Sigma_{\phi}(\x) + \Sigma_{\phi}(\hat{\mathbf{X}}) - 2 \left( \Sigma_{\phi}(\mathbf{X}) \Sigma_{\phi}(\hat{\mathbf{X}}) \right)^{1/2} \right)
\end{align}
where $\mathbf{X}$ and $\hat{\mathbf{X}}$ are the sets of real and reconstructed images, respectively; $\mu_{\phi}(\mathbf{X})$ and $\Sigma_{\phi}(\mathbf{X})$ are the mean and covariance of the features extracted from the real images $\mathbf{X}$ using a pre-trained Inception-v3 model $\phi$ \cite{Szegedy2015Inceptionv3};  $\mu_{\phi}(\hat{\mathbf{X}})$ and $\Sigma_{\phi}(\hat{\mathbf{X}})$ are the corresponding statistics from the reconstructed images $\hat{\mathbf{X}}$. Rather than directly comparing images pixel by pixel, the FID compares the distributions of the latent representations of the original and reconstructed images obtained using a convolutional neural network (Inception v3) trained for image recognition. 
From its definition, FID is a parameter that is robust against translation, rotation, or change of scale of relevant objects. 
A low value of FID indicates that a DNN operating over the two batches of original and reconstructed images extract representations that are statistically equivalent.

\FP{To assess the performance of SQ-GAN, we compare it with established \gls{sota} image encoders, including JPEG2000, BPG, \FP{FCC, and \gls{hific}}, using four metrics: FID, LPIPS, Peak Signal-to-Noise Ratio (PSNR), and mIoU. The first three metrics are computed between the original image $\x$ and its reconstruction $\hat{\x}$, while the \gls{miou} is evaluated between the original semantic segmentation map $\s$ and the \gls{ssm} $\tilde{\s}$, generated from $\hat{\x}$ using the \gls{sota} INTERN-2.5 \gls{ssmodel}. The quantitative results are shown in \fref{fig: SQGAN all metrics comparison}, which plots each metric as a function of BPP, as defined in \eqref{eq: SQGAN BPP}.}

\FP{We begin by examining how SQ-GAN compares to approaches that are able to compress the image at comparable values of BPP, namely JPEG2000, BPG, and FCC. As evident from \fref{fig: SQGAN all metrics comparison}, SQ-GAN significantly outperforms these methods on semantic-aware metrics such as FID, LPIPS, and mIoU, which better reflect perceptual and task-related image quality. SQ-GAN is outperformed by BPG and FCC on PSNR, a pixel-level distortion measure, which favors pixel-by-pixel recovery rather than semantic relevance. }

\FP{
It is important to note that SQ-GAN is capable of operating effectively even at very low bitrates. For instance, even at approximately $0.01$ BPP, it preserves meaningful semantic structures, as demonstrated by favorable FID, LPIPS, and mIoU scores. It is also interesting to observe what happens at BPP $= 0.038$. BPG reconstructs images with a PSNR of 26, outperforming SQ-GAN in terms of pixel fidelity. However, as shown in \fref{fig: SQGAN visual comparison sqgan bpg}, BPG fails to preserve important semantic content. In fact, only the first two cars in the scene are detected while the distant pedestrian and the traffic signs are lost. In contrast, SQ-GAN maintains better object-level consistency.}

\FP{
The comparison with the \gls{fcc} model further highlights the efficiency of SQ-GAN in semantic preservation. For example, FCC achieves an FID=93 at a BPP of $0.061$, whereas SQ-GAN attains the same FID at a BPP of $0.017$, indicating a bitrate reduction factor of approximately $3.6$ times to achieve comparable reconstructed image quality.
Similar trends are observed in LPIPS and mIoU, reinforcing the semantic coding advantage of our model.}

\FP{
Finally, we considered the \gls{hific} model, a High-Fidelity Generative Image Compression method often adopted as a benchmark. We chose not to include HiFiC in \fref{fig: SQGAN all metrics comparison}, as its operational BPP range begins at $0.142$, which is substantially higher than the range analyzed for the other models in the figure (from $0.015$ to $0.08$ BPP).
 Working at BPP $= 0.142$, HiFiC yields superior performance in terms of PSNR$=28.34$, FID$=35.47$, and LPIPS$=0.223$, as expected, because of the much higher BPP. Nevertheless, it is impressive to notice that, in terms of mIoU,  HiFiC reaches only a value of $0.565$, i.e., the same value attained by SQ-GAN at BPP $= 0.018$, which corresponds to a rate reduction factor of about $7.8$ times. This underlines the remarkable semantic compression capability of SQ-GAN, even when compared with high-performing learned codecs at significantly higher bitrates.}

\section{Conclusions}
This work presents \gls{sqgan}, a new approach for simultaneous encoding of images and their semantic maps, aimed at preserving the image semantic content, even at very low coding rates. 
The comparison of \gls{sqgan} with \gls{sota} encoders has been carried out using a set of perceptual performance indicators, like FID, LPIPS, and mIoU, that somehow measure the semantic distortion between the original image (or semantic map) and the reconstructed image (or semantic map). Using these performance indicators, extensive evaluations on the Cityscapes dataset show that
SQ-GAN significantly outperforms JPEG2000, BPG, \FP{FCC and HiFiC,} in terms of rate-(semantic) distortion trade-off. At the same time, SQ-GAN is clearly outperformed by BPG \FP{and FCC} in terms of conventional pixel-level indicators, such as the PSNR. 
\FP{
In summary, SQ-GAN significantly outperforms state-of-the-art image encoders in semantic-aware image compression tasks, whose primary goal is not  perfect pixel-level reconstruction, but rather the preservation of semantic contents that are more important for high-level vision tasks. 
}

\textcolor{black}{We tested our method on the autonomous driving use case,  but our approach can be applied to many other scenarios, like medical imaging, remote sensing, space exploration, etc., whenever there is a need to strongly compress the images, while retaining as much as possible their semantic content.}
\textcolor{black}{
Nevertheless, also the autonomous driving use case  offers useful motivations for using our coding scheme, like reducing the traffic load necessary to send images to a remote server for training autonomous driving models; or strongly compressing images for liability assessment in case of accidents, while preserving critical semantic details (e.g., traffic lights, road signs, pedestrians).}
\textcolor{black}{Further investigations are in progress to simplify the coding strategy, optimize the shapes of the image and semantic map tensors and generalize the approach to video coding.}  
\section*{Acknowledgment}
This work was supported by the Gottfried Wilhelm Leibniz-Preis 2021 of the German Science Foundation (DFG), Joint Project 6G-RIC (Project IDs 16KISK030), the Italian National Recovery and Resilience Plan (NRRP) of NextGenerationEU, partnership on “Telecommunications of the Future” (PE00000001 - program RESTART), Huawei Technology France SASU, under agreement N.TC20220919044, and the SNS-JU-2022 project ADROIT6G under agreement n. 101095363.

\ifCLASSOPTIONcaptionsoff
  \newpage
\fi

\bibliographystyle{IEEEtran}
\bibliography{Bibliography}

\section{Biography Section}
\begin{IEEEbiography}[{\includegraphics[width=1in,height=1.25in,clip,keepaspectratio]{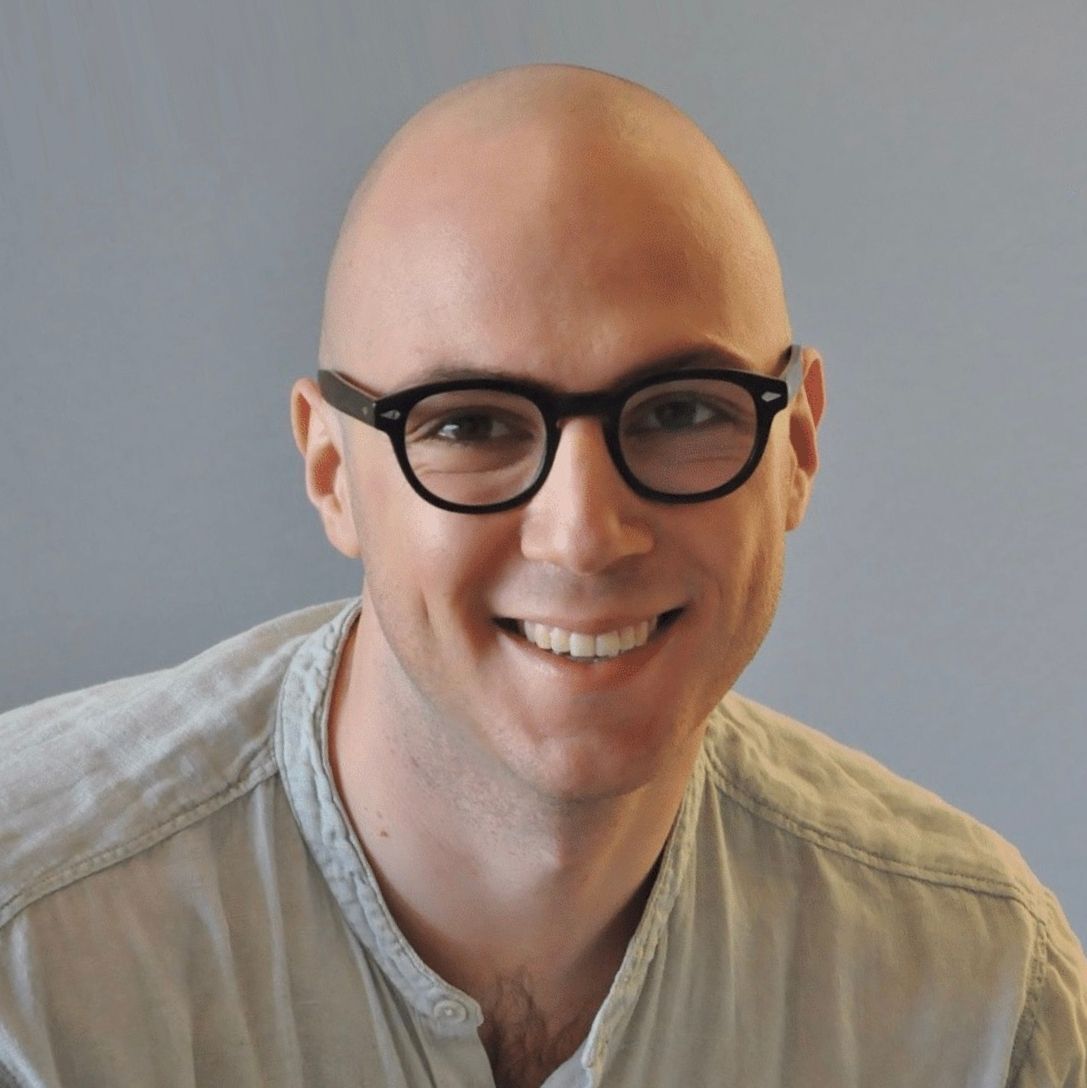}}]{Francesco Pezone} received his B.Sc. in Physics from the University of Rome Tor Vergata, his M.Sc. in Data Science from Sapienza University of Rome, and dual Ph.D. degrees in Data Science from Sapienza University of Rome and Engineering from Technische Universität Berlin. He is an AI researcher specializing in generative AI, computer vision, and optimization theory, with a focus on efficient data compression and semantic communication.  
\end{IEEEbiography} 
\begin{IEEEbiography}[{\includegraphics[width=1in,height=1.25in,clip,keepaspectratio]{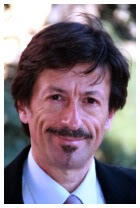}}]{Sergio Barbarossa} is a Full Professor at Sapienza University of Rome and a Senior
Research Fellow of Sapienza School for Advanced Studies (SSAS). He is an IEEE
Fellow and a EURASIP Fellow. He received the Technical Achievements Award
from the European Association for Signal Processing (EURASIP) society in 2010 and
the IEEE Signal Processing Society Best Paper Awards in the years 2000, 2014, and
2020, and. He was an IEEE Distinguished Lecturer in 2013-2014. He has been the
scientific coordinator of several European projects and he is now coordinating a
national project called “Network Intelligence” and a national initiative named “Make
Artificial Intelligence Distributed and Networked”. His main current research
interests include semantic and goal-oriented communications, topological signal
processing and learning, 6G networks and distributed edge machine learning.

\end{IEEEbiography}

\begin{IEEEbiography}[{\includegraphics[width=1in,height=1.25in,clip,keepaspectratio]{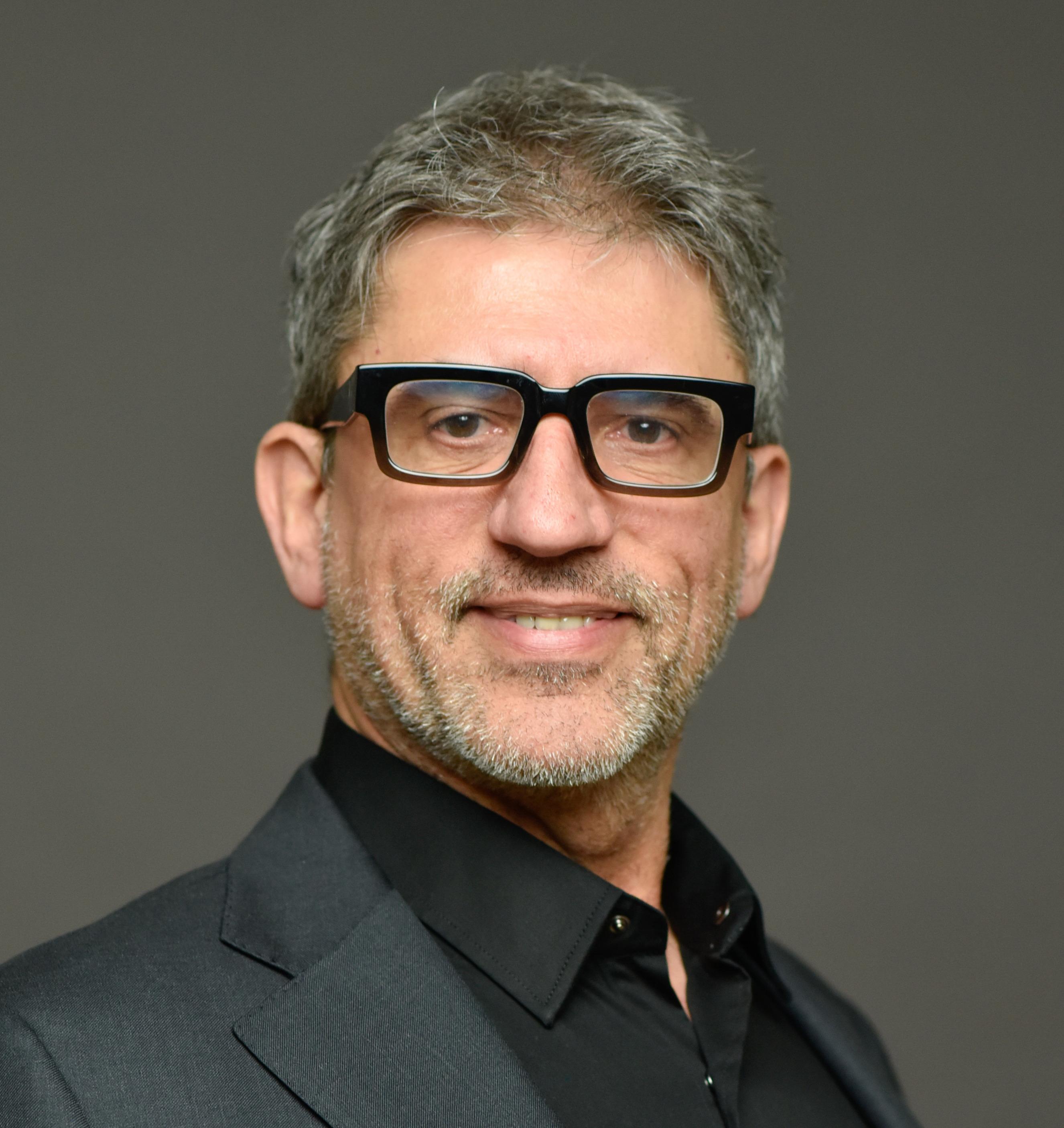}}]{Giuseppe Caire} (S '92 -- M '94 -- SM '03 -- F '05) 
was born in Torino in 1965. He received a
B.Sc. in Electrical Engineering  from Politecnico di Torino in 1990, 
an M.Sc. in Electrical Engineering from Princeton University in 1992, and a Ph.D. from Politecnico di Torino in 1994. 
He has been a post-doctoral research fellow with the European Space Agency (ESTEC, Noordwijk, The Netherlands) in 1994-1995,
Assistant Professor in Telecommunications at the Politecnico di Torino, Associate Professor at the University of Parma, Italy, 
Professor with the Department of Mobile Communications at the Eurecom Institute,  Sophia-Antipolis, France,
a Professor of Electrical Engineering with the Viterbi School of Engineering, University of Southern California, Los Angeles,
and he is currently an Alexander von Humboldt Professor with the Faculty of Electrical Engineering and Computer Science at the Technical University of Berlin, Germany.

He received the Jack Neubauer Best System Paper Award from the IEEE Vehicular Technology Society in 2003,  the
IEEE Communications Society and Information Theory Society Joint Paper Award in 2004, in 2011, and in 2025, 
the Okawa Research Award in 2006,   the Alexander von Humboldt Professorship in 2014, the Vodafone Innovation Prize in 2015, an ERC Advanced Grant in 2018,  the Leonard G. Abraham Prize for best IEEE JSAC paper in 2019, the IEEE Communications Society Edwin Howard Armstrong Achievement Award in 2020, the 2021 Leibniz Prize  of the German National Science Foundation (DFG), and the  CTTC Technical Achievement Award of the IEEE Communications Society in 2023.  Giuseppe Caire is a Fellow of IEEE since 2005.  He has served in the Board of Governors of the IEEE Information Theory Society from 2004 to 2007, and as officer from 2008 to 2013. He was President of the IEEE Information Theory Society in 2011. 
His main research interests are in the field of communications theory, information theory, channel and source coding
with particular focus on wireless communications.   
\end{IEEEbiography}

\end{document}